\pdfoutput=1

\documentclass[11pt]{article}
\usepackage{amsthm}

\usepackage[preprint]{acl}

\usepackage{times}
\usepackage{latexsym}

\usepackage[T1]{fontenc}

\usepackage[utf8]{inputenc}

\usepackage{microtype}

\usepackage{inconsolata}

\usepackage{graphicx}
\usepackage{amsmath}
\usepackage{algorithm}
\usepackage{algorithmic}
\usepackage{multirow}
\usepackage{booktabs}
\usepackage{amsfonts}
\usepackage{newfloat}
\usepackage{listings}

\newtheorem{theorem}{Theorem}
\newtheorem{corollary}{Corollary}
\title{Rethinking DPO: The Role of Rejected Responses in Preference Misalignment}

\author{Jay Hyeon Cho, JunHyeok Oh, Myunsoo Kim, Byung-Jun Lee \\
        Korea University\\
        \texttt{\{bonin147, the2nlaw, m970326, byungjunlee\}@korea.ac.kr}}

\begin{document}
\maketitle
\begin{abstract}
Direct Preference Optimization (DPO) is a simple and efficient framework that has attracted substantial attention. However, it often struggles to meet its primary objectives---increasing the generation probability of chosen responses while reducing that of rejected responses---due to the dominant influence of rejected responses on the loss function. This imbalance leads to suboptimal performance in promoting preferred responses. In this work, we systematically analyze the limitations of DPO and existing algorithms designed to achieve the objectives stated above. To address these limitations, we propose Bounded-DPO (BDPO), a novel method that bounds the influence of rejected responses while maintaining the original optimization structure of DPO. Through theoretical analysis and empirical evaluations, we demonstrate that BDPO achieves a balanced optimization of the chosen and rejected responses, outperforming existing algorithms.
\end{abstract}

\section{Introduction}

Aligning Large Language Models (LLMs) with human values using human preference data has become essential in natural language processing (NLP). Direct Preference Optimization (DPO; \citealt{rafailov2024direct}) has gained significant attention for efficiently optimizing preference comparisons directly from pairwise feedback, without relying on explicit reward models or reinforcement learning, offering a computationally cheaper alternative to conventional RLHF methods~\citep{ouyang2022training, bai2022training, stiennon2020learning}.

Despite its simplicity, DPO exhibits two key limitations. First, it tends to increase the probability of generating out-of-distribution (OOD) actions, a drawback attributed to the absence of sampling during training~\citep{xudpo}. Second, several studies~\citep{pang2024iterative, adler2024nemotron, liu2024provably} have highlighted that DPO struggles to sufficiently increase the probability of chosen actions. 
To address this, DPO+NLL was proposed, augmenting DPO with a negative log-likelihood (NLL) loss—commonly used in supervised learning—to explicitly improve the likelihood of chosen responses. Another alternative, DPO-Positive (DPOP), introduces a penalty when the probability of a chosen response falls below that assigned by the initial model~\citep{pal2024smaug}.

This study identifies two core objectives of DPO: (1) increasing the generation probability of chosen responses while decreasing that of rejected responses, and (2) ensuring the model does not deviate too far from the reference model. 
The analysis reveals that, in theory, the DPO loss can achieve an optimal solution by simply lowering the probability of generating rejected responses, without necessarily increasing the probability of generating chosen responses. Empirical results corroborate this finding, showing that DPO fails to adequately fulfill objective (1).
This work also examines alternative approaches, such as DPO+NLL and DPOP, and discusses their respective limitations in detail.

We propose Bounded-DPO (BDPO), a novel algorithm designed to address the two primary objectives of DPO more effectively. BDPO preserves the overall structure of the original DPO loss, but replaces the rejected response distribution of the updated model with a mixture distribution that incorporates the reference model. This modification bounds the influence of rejected responses on the loss function, ensuring adherence to objective (2) while effectively achieving objective (1).

Following the introduction of BDPO, we present a theoretical analysis showing that the BDPO loss guarantees a lower bound on the probability of generating the chosen response (Section~\ref{sec:Bounded-DPO}). We conduct both toy examples and real-world model experiments to compare the behavior of BDPO with existing algorithms (Section~\ref{sec:Experimental Validation of BDPO}). Experimental results demonstrate that BDPO achieves superior performance in preference optimization (Section~\ref{sec:Experiment}). Additionally, we conducted an ablation study to gain deeper insights into BDPO's performance and provided a detailed analysis of the findings.

\section{Related Works}
Direct Preference Optimization (DPO, \citealt{rafailov2024direct}) simplifies the RLHF framework by eliminating the need for reward modeling, instead directly training models on human preference data. Due to its efficiency and simplicity, DPO has been widely adopted in various NLP applications~\citep{dubey2024llama, yuanself, chen2024self}.

However, despite its practical advantages, DPO often yields models that underperform compared to those trained with conventional RLHF methods~\citep{ivison2024unpacking, xudpo, tang2024understanding}. Recent studies have identified several notable limitations of DPO, including:
\begin{itemize}
\item susceptibility to reward divergence, which can lead to overfitting~\citep{azar2024general}
\item 
a tendency to learn a biased policy that favors OOD responses~\citep{xudpo}

\item difficulty in correcting inaccurate preference rankings~\citep{chen2024preference}
\item failure to promote chosen responses effectively during optimization~\citep{pal2024smaug, rafailov2024r, pang2024iterative, razin2024unintentional, tajwarpreference, liu2024provably, xie2024minor}
\end{itemize}

Among these, we consider the last one to be the most significant and aim to address it in this paper, as it fundamentally undermines the core objective of DPO.\footnote{These issues are often interrelated; for example, reducing the probabilities of both chosen and rejected responses may increase the likelihood of generating OOD responses.} To the best of our knowledge, only two prior approaches have addressed this problem without relying on online sampling during training. We provide an in-depth comparison and analysis of these methods in this work.

\paragraph{DPO+NLL} To prevent DPO from reducing the probability of chosen responses, a simultaneous supervised learning step on chosen responses has been proposed~\citep{adler2024nemotron, pang2024iterative, liu2024provably}. This method, referred to as DPO+NLL in this paper, adds an explicit negative log-likelihood (NLL) loss term to the original DPO objective to directly increase the likelihood of chosen responses. However, we show that it introduces sensitivity to hyperparameters, often causing the training process to be dominated by the NLL term. 

\paragraph{DPO-Positive (DPOP)} 
Another approach introduces direct penalties when DPO fails to increase the probability of chosen responses to guide the optimization process~\citep{pal2024smaug}. However, we show that this method does not overcome the fundamental limitations of DPO and ultimately exhibits similar shortcomings.

\begin{figure*}[t!]
\centerline{\includegraphics[width=2\columnwidth]{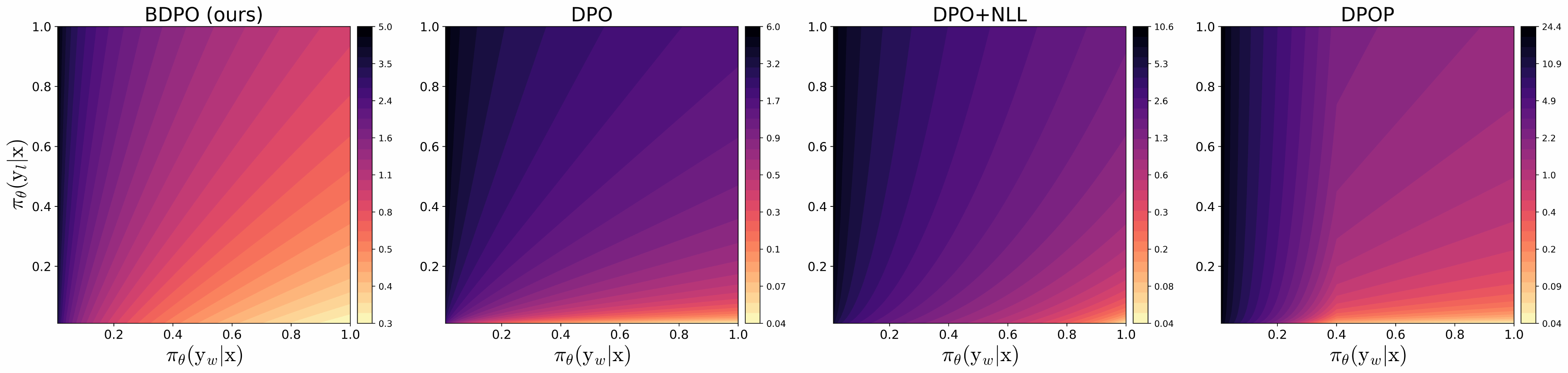}}
  \caption{Contour maps of the loss with reference model \(\pi_\mathrm{ref}(\mathbf{y}_w|\mathbf{x}) = 0.4\) and \(\pi_\mathrm{ref}(\mathbf{y}_l|\mathbf{x}) = 0.1\). 
  DPO shows that having \(\pi_\theta(\mathbf{y}_l|\mathbf{x})\) close to zero results in the lowest loss, regardless of \(\pi_\theta(\mathbf{y}_w|\mathbf{x})\). DPOP applies penalties below \(\pi_\mathrm{ref}(\mathbf{y}_w|\mathbf{x})\) but shares the same issue to DPO. 
  In contrast, BDPO and DPO+NLL are not dominated by \(\pi_\theta(\mathbf{y}_l|\mathbf{x})\), and they are able to effectively learn to increase \(\pi_\theta(\mathbf{y}_w|\mathbf{x})\).}
  \label{fig:heatmap}
\end{figure*}

\section{Analysis of Existing Objectives for Preference Optimization}
In this section, we revisit the objectives of DPO and review prior findings showing that it can deviate from its intended goals.
To deepen the understanding of DPO's behavior, we provide a detailed analysis of its underlying mechanisms.
We also examine the limitations of two representative extensions—DPO+NLL and DPOP—that have been proposed to mitigate these issues.


The DPO objective is defined as:
\begin{align}
\label{eq:DPO_1}
\max_\theta~\mathbb{E}_{\mathcal{D}}\left[\log\sigma\left(r_\theta(\mathbf{y}_w) - r_\theta(\mathbf{y}_l)\right)\right],
\end{align}
where \(r_\theta(\mathbf{y}) := \beta\log \frac{\pi_\theta(\mathbf{y}|\mathbf{x})}{\pi_\mathrm{ref}(\mathbf{y}|\mathbf{x})}\), \(\sigma\) is the sigmoid function, \(\pi_{\theta}\) denotes the learned policy, \(\pi_{\mathrm{ref}}\) is the reference model, and \(\beta\) is a hyperparameter.

The DPO loss captures two key objectives:
\begin{itemize}
    \item \textbf{Optimizing response probabilities:} Increase the probability of the chosen response while decreasing that of the rejected response.
    \item \textbf{Alignment with the reference model:} Ensure the learned model does not deviate significantly from the reference model.
\end{itemize}
These characteristics mirror those of the original RLHF framework. The first reflects the use of reward models trained to favor chosen responses over rejected ones, with policies optimized to maximize these rewards. The second aligns with behavior regularization, which constrains the learned policy to remain close to the reference model and mitigates over-optimization on uncertain rewards.
\subsection{Does DPO Increase the Probability of the Chosen Response?}
\label{sec:Does DPO Increase Chosen Probability?}
\citet{xudpo} demonstrated through illustrative experiments that, under DPO, the probabilities of both the chosen and rejected responses can simultaneously decrease. They further highlighted that DPO is prone to incorrect learning behavior, particularly in OOD scenarios. To provide a more intuitive perspective on the DPO objective, the loss can be reformulated as:
\begin{equation}
\label{eq:DPO_2}
\max_\theta~\mathbb{E}_{\mathcal{D}}\left[\log \sigma\left(\hat{s}_\theta(w; l) - \hat{s}_{\mathrm{ref}}(w; l)\right)\right],
\end{equation}
where \(\hat{s}_\theta(w; l) := \beta\log \frac{\pi_\theta(\mathbf{y}_w|\mathbf{x})}{\pi_\theta(\mathbf{y}_l|\mathbf{x})}\) and \(\hat{s}_\mathrm{ref}(w; l) := \beta\log \frac{\pi_\mathrm{ref}(\mathbf{y}_w|\mathbf{x})}{\pi_\mathrm{ref}(\mathbf{y}_l|\mathbf{x})}\) denote the scaled log-ratios under the learned policy \(\pi_\theta\) and the reference model \(\pi_\mathrm{ref}\).

Both the logarithm and sigmoid functions are strictly increasing, and \(\pi_\mathrm{ref}\) remains fixed during training. As a result, minimizing the DPO loss is equivalent to maximizing the term \(\hat{s}_\theta(w;l)\).
A straightforward way to achieve this maximization is by reducing \(\pi_{\theta}(\mathbf{y}_{l}|\mathbf{x})\) to zero. However, setting \(\pi_{\theta}(\mathbf{y}_{l}|\mathbf{x})=0\) leads to an unbounded loss, and the objective can be optimized regardless of the probability assigned to the chosen response \(\pi_{\theta}(\mathbf{y}_{w}|\mathbf{x})\).


Furthermore, even when training proceeds as intended—i.e., increasing the log-ratio for the chosen response while decreasing it for the rejected one—the concavity of the logarithm disproportionately amplifies the effect of the rejected response. As a result, the optimization may become dominated by the rejected term.

\paragraph{Loss Visualization} Figure~\ref{fig:heatmap} illustrates how the loss varies with respect to the chosen response probability \(\pi_\theta(\mathbf{y}_w|\mathbf{x})\) and the rejected probability \(\pi_\theta(\mathbf{y}_l|\mathbf{x})\). 
Contour lines represent regions of equal loss, with brighter colors indicating lower values. During training, optimization tends to move toward these brighter (lower-loss) regions.
The DPO subplot clearly shows that minimizing \(\pi_\theta(\mathbf{y}_l|\mathbf{x})\) alone yields the lowest loss, regardless of the value of \(\pi_\theta(\mathbf{y}_w|\mathbf{x})\), highlighting a strong dependence on the rejected response probability. We empirically confirm that this behavior occurs in practice by analyzing the convergence patterns of DPO loss in a real-world language model (Figure~\ref{fig:Learning_pattern}, Section~\ref{sec:Training Dynamics of BDPO}).



\paragraph{Remark}
Given a prompt \(\mathbf{x}\), the model assigns probabilities over all possible responses such that they sum to one. 
If the expected update to both the chosen and rejected responses is negative, i.e., $\mathbb{E}_{\mathcal{D}}[\Delta\pi_{\theta}(\mathbf{y}_{w}|\mathbf{x})+\Delta\pi_{\theta}(\mathbf{y}_{l}|\mathbf{x})]<0$, the probability mass is implicitly shifted toward out-of-distribution (OOD) responses. We empirically confirm this shift toward OOD responses in Appendix~\ref{sec:Out-of-Distribution (OOD) Probability Analysis}.

\subsection{Analysis of DPO+NLL Loss}
\label{sec:Analysis of NLL Loss}
In supervised fine-tuning, the model is typically trained to minimize the Negative Log-Likelihood (NLL) loss, which is defined as: 
\begin{equation*}
\mathcal{L}_{\mathrm{NLL}}(\pi_\theta)=-\mathbb{E}_{\mathcal{D}}\left[\log\pi_\theta(\mathbf{y}|\mathbf{x})\right],
\end{equation*}
where \(\mathbf{y}\) denotes the target response. Minimizing this objective encourages the model to assign higher probability to \(\mathbf{y}\).

To mitigate the issue of decreasing probabilities for the chosen response in DPO, \citet{pang2024iterative, adler2024nemotron, liu2024provably} introduced augmenting the DPO loss with the NLL loss, using the chosen response \(\mathbf{y}_w\) as the target label for NLL loss. The resulting DPO+NLL loss is:
\begin{equation*}
\mathcal{L}_{\mathrm{DPO+NLL}}=\mathcal{L}_{\mathrm{DPO}}+\alpha\mathcal{L}_{\mathrm{NLL}},
\end{equation*}
where $\alpha$ is a hyperparameter controlling the relative weight of the NLL term.

Since the NLL loss explicitly increases the probability of the chosen response, this objective can address the limitations of DPO when $\alpha$ is sufficiently large. Moreover, the NLL term can be viewed as the forward KL divergence between the chosen response distribution and the updated model, offering an intuitive explanation of how it mitigates DPO's vulnerability to OOD responses.

However, we show that when $\alpha$ is large enough to fully correct DPO's shortcomings, \textbf{the update process becomes dominated by the NLL loss}. This dominance can cause the updated model to deviate significantly from the reference model.



Examining the gradients of each component in the DPO+NLL objective, we obtain:

\begin{align*}
&\nabla_\theta\mathcal{L}_{\mathrm{DPO}}=-\mathbb{E}_{\mathcal{D}}\bigg[
\beta\cdot\sigma\left(r_\theta(\mathrm{y}_{l})-r_\theta(\mathrm{y}_{w})\right)\\\nonumber
&~~~~~~~~~~~\cdot\left[\nabla_\theta\log\pi_\theta(\mathbf{y}_w|\mathbf{x})-\nabla_\theta\log\pi_\theta(\mathbf{y}_l|\mathbf{x})\right]\bigg],\\
&\nabla_\theta\mathcal{L}_{\mathrm{NLL}}=-\mathbb{E}_{\mathcal{D}}[\nabla_\theta\log\pi_\theta(\mathbf{y}_w|\mathbf{x})].
\end{align*}
From this, we observe that the total gradient with respect to \(\log\pi_\theta(\mathrm{y}_w|\mathrm{x})\) is scaled by:
\begin{equation*}
    \beta\cdot\sigma\left(r_\theta(\mathrm{y}_{l})-r_\theta(\mathrm{y}_{w})\right)+\alpha.
\end{equation*}


This coefficient determines the extent to which the probability of the chosen response $\mathbf{y}_w$ is increased during training.
As training progresses, the value of  \(r_\theta(\mathrm{y}_{l}|\mathrm{x})-r_\theta(\mathrm{y}_{w}|\mathrm{x})\) tends to \(-\infty\), causing the sigmoid \(\sigma(\cdot)\) to approach zero. In contrast, $\alpha$ remains a fixed constant, which results in the NLL term increasingly dominating the update.


Moreover, commonly used hyperparameter settings (e.g., $\beta = 0.1$, $\alpha = 1$) imply that the contribution of the NLL component to the gradient of $\mathbf{y}_w$ outweighs that of the DPO component by at least a factor of 10. This further supports the concern that DPO+NLL tends to move the model away from the reference distribution, contrary to DPO’s original intent. Additionally, if the NLL term dominates optimization, the risk of overfitting increases. This trade-off between correcting DPO’s limitations and preserving alignment with the reference model makes the choice of $\alpha$ highly nontrivial, as illustrated in the following analysis.

\begin{figure}[htbp]
  \centerline{\includegraphics[width=\columnwidth]{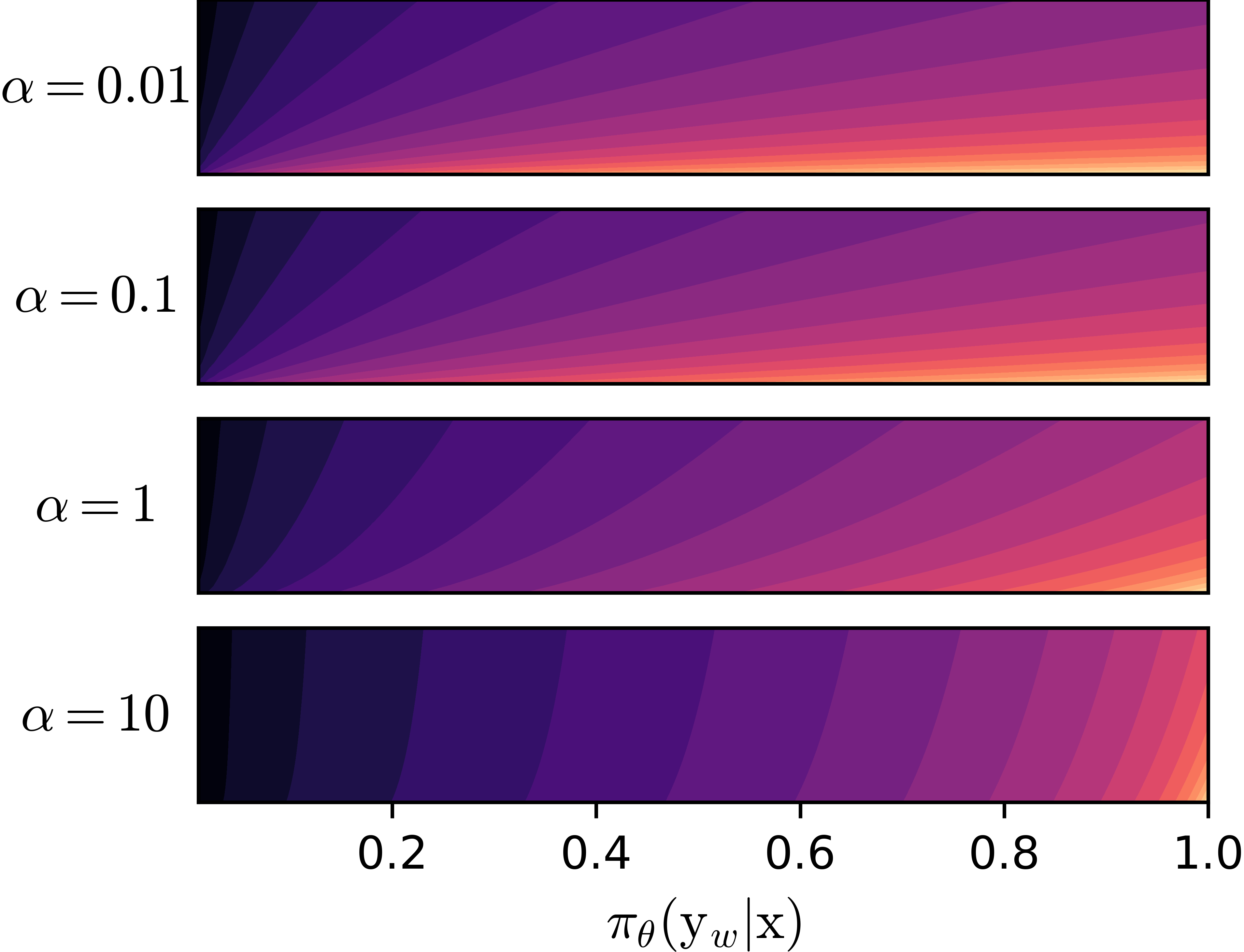}}
  \caption{Contour maps for various values of $\alpha$ $\in$ $\{$$0.01$, $0.1$, $1$, $10$$\}$, focusing on the region of interest where \(\pi_\theta(\mathbf{y}_l|\mathbf{x})\in [0, 0.25]\). The results show that, depending on the value of \(\alpha\), DPO+NLL may either fail to address the limitations of DPO (low $\alpha$), or cause significant deviation from the reference model (high $\alpha$).}
  \label{fig:dpo_nll_heatmaps}
\end{figure}

\paragraph{Loss Visualization} Figure~\ref{fig:dpo_nll_heatmaps} extends the loss visualization from Figure~\ref{fig:heatmap} by presenting contour maps of the DPO+NLL loss for different values of $\alpha$. For \(\alpha = 0.01\) and \(\alpha = 0.1\), the behavior remains consistent with that of DPO, failing to address the limitations. At \(\alpha = 1\), the optimization demonstrates desirable behavior, aligning well with the intended learning dynamics. However, when \(\alpha = 10\), the optimization process exhibits extreme loss patterns, which leads the updated model to significantly deviate from the reference model \(\pi_\mathrm{ref}(\mathbf{y}_w|\mathbf{x}) = 0.4\) and \(\pi_\mathrm{ref}(\mathbf{y}_l|\mathbf{x}) = 0.1\). This highlights the sensitivity of the algorithm to \(\alpha\).

\subsection{Analysis of DPOP Loss}
\citet{pal2024smaug} proposed DPO-Positive (DPOP) to address the limitations of DPO by introducing a penalty term. The DPOP objective is defined as:
\begin{align*}
& \max_\theta\mathbb{E}_{\mathcal{D}} \big[ 
 \log \sigma \big( r_\theta(\mathbf{y}_{w})-r_\theta(\mathbf{y}_{l})
\nonumber\\&~~~~~~~~~~~~~~~~~~ - \lambda \cdot \max \left( 0, -r_\theta\left(\mathbf{y}_{w}\right) \right) 
\big)\big],
\end{align*}
where $\lambda > 0$ is a hyperparameter. The penalty is activated when the reward for the chosen response \(r_\theta(\mathrm{y}_{w})\) falls below zero, which corresponds to the case where \( \log \frac{\pi_{\mathrm{ref}}(\mathbf{y}_{w}|\mathbf{x})}{\pi_{\theta}(\mathbf{y}_{w}|\mathbf{x})} > 0 \).

Figure~\ref{fig:heatmap} shows that when \(\pi_\theta(\mathbf{y}_w|\mathbf{x}) > \pi_\mathrm{ref}(\mathbf{y}_w|\mathbf{x})\), DPOP behaves identically to DPO, whereas for \(\pi_\theta(\mathbf{y}_w|\mathbf{x}) < \pi_\mathrm{ref}(\mathbf{y}_w|\mathbf{x})\), significantly higher loss levels are observed, clearly indicating the region where the penalty is applied. Nevertheless, as the penalty term becomes substantial only when \(\pi_\theta(\mathbf{y}_{w}|\mathbf{x})\) approaches zero, the limitations of DPO are not fully addressed. The DPOP loss still guarantees an optimal solution where $\pi_\theta(\mathbf{y}_{l}|\mathbf{x}) = 0$ and $\pi_\theta(\mathbf{y}_{w}|\mathbf{x})\neq 0$.
As discussed in Section~\ref{sec:Does DPO Increase Chosen Probability?}, the steep divergence of the logarithmic function near zero causes disproportionate impacts when probabilities are low. Although the penalty term in DPOP aims to stabilize optimization by discouraging undesirable updates, it fails to fully mitigate the inherent imbalance introduced by this property. In Section~\ref{sec:Examining the OOD Behavior}, we demonstrate that DPOP, like DPO, is heavily influenced by rejected responses, which can result in an increased probability of generating OOD responses.


\section{Our Algorithm}
In this section, we introduce Bounded-DPO (BDPO), our proposed method designed to satisfy the two primary objectives of DPO. We provide its theoretical foundation and show empirically that BDPO achieves the intended behavior while overcoming the limitations of existing methods.
\subsection{Bounded-DPO}
\label{sec:Bounded-DPO}
Bounded-DPO (BDPO) extends the DPO formulation in Equation~\eqref{eq:DPO_2} by replacing \(\pi_\theta(\mathbf{y}_{l}|\mathbf{x})\) with a mixture distribution.
The BDPO loss is defined as:
\begin{align*}
&\mathcal{L}_{\mathrm{BDPO}}(\pi_{\theta};\pi_{\mathrm{ref}}):=-\mathbb{E}_{\mathcal{D}}\\&~~~~~~~~~~\left[\log\sigma\left(\beta\log\frac{\pi_{\theta}(\mathbf{y}_{w}|\mathbf{x})}{\pi_{\mathrm{mix}}(\mathbf{y}_{l}|\mathbf{x})}-\hat{s}_\mathrm{ref}(w;l)\right)\right],\nonumber
\end{align*}
where \(\pi_\mathrm{mix}(\mathbf{y}|\mathbf{x})=\lambda\pi_{\theta}(\mathbf{y}|\mathbf{x})+(1-\lambda)\pi_{\mathrm{ref}}(\mathbf{y}|\mathbf{x}),\) for \(\lambda \in (0, 1)\). 

This formulation ensures that the denominator is lower bounded by \((1-\lambda)\pi_{\mathrm{ref}}(\mathbf{y}_l|\mathbf{x})\), even when \(\pi_{\theta}(\mathbf{y}_{l}|\mathbf{x})=0\). This property is essential for ensuring that the loss contributes to the intended learning behavior. The parameter $\lambda$ controls the balance between the learned policy and the reference model: larger values of $\lambda$ give more weight to $\pi_\theta$, while smaller values emphasize $\pi_{\mathrm{ref}}$. The effect of $\lambda$ is further analyzed in Section~\ref{sec:Ablation study}.

To analyze the behavior of BDPO, we consider the simplest scenario with a single response pair in the dataset, $\mathcal{D}=\{(\mathbf{y}_w, \mathbf{y}_l)\}$.\footnote{We do not consider the case where multiple response pairs are associated with a single prompt.}
Under this condition, BDPO exhibits the following properties (proofs are deferred to the Appendix~\ref{proof:converge}):
\begin{theorem}
\label{thm:converge}
Let \(\pi^*\) denote the policy that minimizes $\mathcal{L}_{\mathrm{BDPO}}$. If \(\pi_{\mathrm{ref}}(\mathbf{y}_{l}|\mathbf{x}) > 0\), then \(\pi^*\) satisfies the following conditions:\\
\[\pi^*(\mathbf{y}_{w}|\mathbf{x})=1\quad \text{and}\quad \pi^*(\mathbf{y}_{l}|\mathbf{x})=0.\]
\end{theorem}
\begin{corollary}
\label{cor:BDPO-dpo-optimality}
Let \(\pi^*\) denote the policy that minimizes $\mathcal{L}_{\mathrm{BDPO}}$. Then \(\pi^*\) also minimizes $\mathcal{L}_{\mathrm{DPO}}$. However, the converse does not hold. 
\end{corollary}

\begin{figure*}[t!]
  \centerline{\includegraphics[width=2\columnwidth]{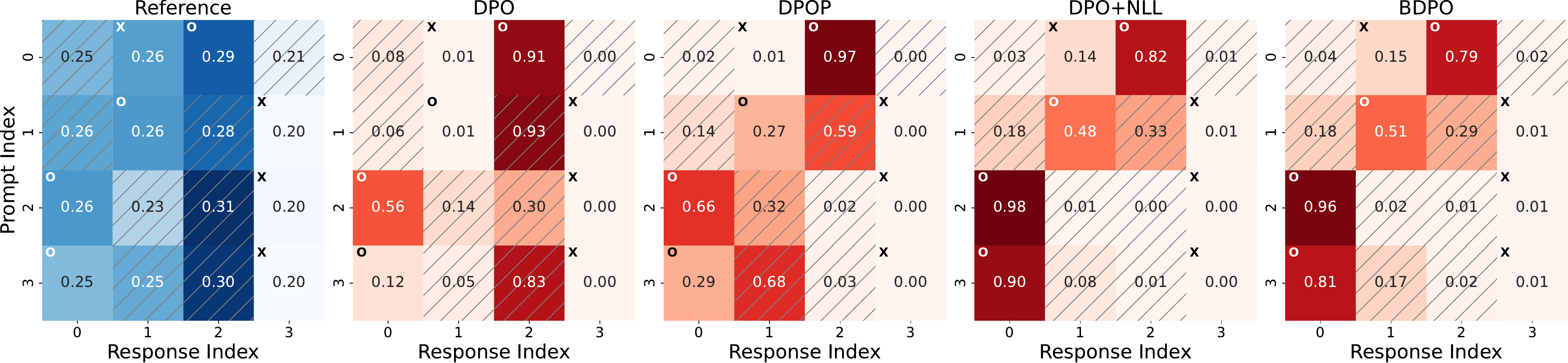}}
  \caption{Training behavior of DPO, DPOP, DPO+NLL, and BDPO on a toy task with four prompts and four responses per prompt. Two out-of-distribution (OOD) responses per prompt are shaded in gray with diagonal stripes. Chosen and rejected responses are marked with o and ×, respectively. While DPO and DPOP fail to suppress some OOD responses, BDPO and DPO+NLL demonstrate more appropriate learning behavior.}
  \label{fig:Toy_main}
\end{figure*}

As discussed in Section~\ref{sec:Does DPO Increase Chosen Probability?}, 
\(\pi_\theta(\mathbf{y}_{l}|\mathbf{x})\) approaching 0 can trivially satisfy the optimality conditions of the DPO loss. In contrast, BDPO addresses this issue by introducing a lower bound in the denominator, ensuring that the problem does not arise even when \(\pi_\theta(\mathbf{y}_{l}|\mathbf{x})\rightarrow0\). This robustness is formally presented in  Theorem~\ref{thm:converge} and Corollary~\ref{cor:BDPO-dpo-optimality}, which show that BDPO avoids this core limitation of DPO.
Additionally, BDPO provides a lower bound for the chosen probability, as stated in the following theorem (proof is deferred to the Appendix~\ref{proof:thm:chosen probabilty lower bound}):
\begin{theorem}
\label{thm:chosen probabilty lower bound}
Let the initial point of \(\pi_\theta\) be \(\pi_\mathrm{ref}\). Suppose that every optimization step ensures the BDPO loss decreases monotonically.  
Then, for any step, \(\pi_\theta\) satisfies the following condition:
\[(1-\lambda)\pi_\mathrm{ref}(\mathbf{y}_w|\mathbf{x})\leq\pi_\theta(\mathbf{y}_w|\mathbf{x})\]
\end{theorem}




\paragraph{Loss Visualization} Figure~\ref{fig:heatmap} also illustrates the loss of the BDPO objective. Unlike DPO and DPOP, BDPO exhibits its minimum loss near \(\pi_\theta(\mathbf{y}_w|\mathbf{x}) = 1\) and \(\pi_\theta(\mathbf{y}_l|\mathbf{x}) = 0\), clearly guiding learning in the desired direction. This result is consistent with the theoretical guarantees provided in Theorem~\ref{thm:converge}.

To further analyze BDPO, we compute the gradient of its loss with respect to \(\pi_\theta(\mathbf{y}_l|\mathbf{x})\):
\begin{align*}
\nabla_{\pi_\theta(\mathbf{y}_l|\mathbf{x})} \mathcal{L}&_{\mathrm{BDPO}}(\pi_{\theta};\pi_{\mathrm{ref}})=
\\&\mathbb{E}_{\mathcal{D}}\left[\beta\sigma\left(-\Delta_\mathrm{BDPO}\right)\cdot\left[\frac{\lambda}{\pi_\mathrm{mix}(\mathbf{y}_l|\mathbf{x})}\right]\right]\nonumber,
\end{align*}
where \(\Delta_\mathrm{BDPO}=\beta\log\frac{\pi_{\theta}(\mathbf{y}_{w}|\mathbf{x})}{\pi_{\mathrm{mix}}(\mathbf{y}_{l}|\mathbf{x})}-\hat{s}_\mathrm{ref}(w;l)\).
In contrast, the gradient of the DPO loss is:
\begin{align*}
\nabla_{\pi_\theta(\mathbf{y}_l|\mathbf{x})} \mathcal{L}&_{\mathrm{DPO}}(\pi_{\theta};\pi_{\mathrm{ref}})=
\\&\mathbb{E}_{\mathcal{D}}\left[\beta\sigma\left(-\Delta_\mathrm{DPO}\right)\cdot\left[\frac{1}{\pi_\theta(\mathbf{y}_l|\mathbf{x})}\right]\right]\nonumber,
\end{align*}
where \(\Delta_\mathrm{DPO}=r_\theta(\mathbf{y}_w) - r_\theta(\mathbf{y}_l)\).
In case of other losses, DPO+NLL shares the identical partial derivatives, and DPOP follows a similar structure (see Appendix~\ref{sec:Details on the Derivative} for details).
From these, we can observe the following:
\begin{itemize}
    \item DPO, DPO+NLL, and DPOP: The gradient term \(\frac{1}{\pi_\theta(\mathbf{y}_l|\mathbf{x})}\) leads to instability as \(\pi_\theta(\mathbf{y}_l|\mathbf{x}) \to 0\), causing unbounded updates and numerical issues.
    \item BDPO: The mixture term \(\pi_\mathrm{mix}(\mathbf{y}_l|\mathbf{x})\) provides a lower bound, preventing divergence even when \(\pi_\theta(\mathbf{y}_l|\mathbf{x}) \to 0\), thereby ensuring stable optimization.  
\end{itemize}









\subsection{Empirical Validation of BDPO}
\label{sec:Experimental Validation of BDPO}
We empirically evaluate whether BDPO and prior methods fulfill the two core objectives of DPO. First, in a toy example (Section~\ref{sec:Examining the OOD Behavior}), we qualitatively compare different loss functions. The results show that both DPO and DPOP are prone to generating OOD actions. Second, using a real-world language model and dataset (Section~\ref{sec:Training Dynamics of BDPO}), we analyze learning dynamics. We find that BDPO behaves as intended, whereas DPO+NLL fails to maintain alignment with the reference model, violating the second objective of DPO.

\subsubsection{Examining the OOD Behavior}
\label{sec:Examining the OOD Behavior}
We construct a toy scenario with four prompts, each paired with four candidate responses. For each prompt, one pair of preference data is randomly selected, and the remaining two responses are implicitly treated as OOD. Further details on the experimental setup are provided in Appendix~\ref{sec:Experimental Details for the Toy Case}.

As shown in Figure~\ref{fig:Toy_main}, DPO and DPOP exhibit undesirable behavior for OOD responses, particularly in prompts 1 and 3. While DPOP prevents a decrease in the probability of the chosen response due to its explicit penalty term, it fails to increase it meaningfully—in prompt 1, the probability remains nearly identical to that of the reference model. Both DPO and DPOP tend to focus primarily on reducing the probability of the rejected response. In contrast, BDPO and DPO+NLL demonstrate desirable learning behavior. The learning dynamics of the toy case training process are provided in Appendix~\ref{sec:Learning Dynamics for the Toy Case}.\footnote{Due to the extreme setup, models may diverge further from the reference model than typically observed.}

\begin{figure*}[t!]
  \centerline{\includegraphics[width=2\columnwidth]{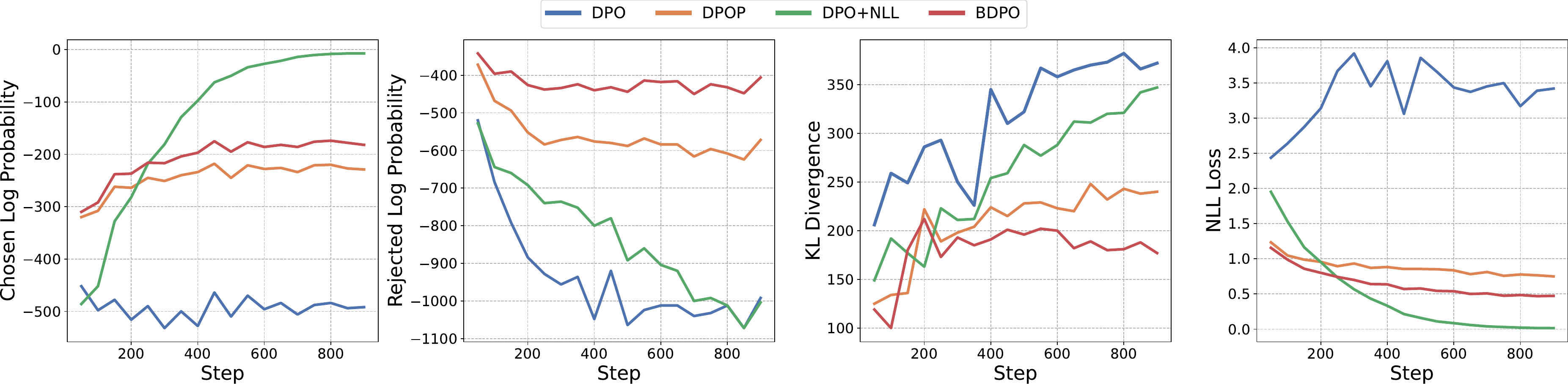}}
  \caption{
Learning dynamics of four algorithms—DPO, DPOP, DPO+NLL, and BDPO—are shown. All algorithms, except DPO, exhibit the desirable behavior of increasing the probability of chosen responses while decreasing the probability of rejected responses. However, DPO+NLL exhibits significant deviation from the reference model, as shown by both the log probabilities and the KL divergence. Although BDPO does not include an explicit NLL term, it achieves a reduction in the NLL loss on the training set.
  }
  \label{fig:Learning_pattern}
  \vspace{-0.3cm}
\end{figure*}

\subsubsection{Training Dynamics of BDPO}
\label{sec:Training Dynamics of BDPO}

We trained the QWEN 0.5B model~\cite{yang2024qwen2} using the UltraFeedback Binarized dataset~\cite{cui2024ultrafeedback}. To better analyze training dynamics, we subsampled the dataset to 1\% of its original size and trained the model for 100 epochs. Additional experimental details are provided in Appendix~\ref{sec:Experimental Details for the Real-World Language Model}.

Figure~\ref{fig:Learning_pattern} presents the log probabilities of the chosen and rejected responses during training. Under DPO, both values decrease, with the rejected probability dropping more sharply—consistent with our analysis in Section~\ref{sec:Does DPO Increase Chosen Probability?}. As a result, the model likely shifts probability mass toward OOD responses, a phenomenon further analyzed in Appendix~\ref{sec:Out-of-Distribution (OOD) Probability Analysis}.

In contrast, BDPO, DPOP, and DPO+NLL exhibited more desirable training dynamics, with the probability of the chosen response increasing and the probability of the rejected response decreasing. Among these, BDPO and DPOP showed relatively smaller changes in their log probabilities compared to their initial values, maintaining closer alignment with the reference model. On the other hand, DPO+NLL pushes the chosen probability toward 100\% and the rejected probability toward 0\%, indicating over-optimization and divergence from the reference. This trend is further confirmed by the KL divergence measurements: both DPO and DPO+NLL exhibit significant deviation from the reference model, as discussed in Sections~\ref{sec:Does DPO Increase Chosen Probability?} and~\ref{sec:Analysis of NLL Loss}, while BDPO and DPOP remain closely aligned. Qualitative results in Appendix~\ref{sec:Full Responses from the Qualitative Analysis} further support these findings.

Interestingly, although BDPO does not include an explicit NLL term, it still achieves a consistent reduction in NLL loss during training. This suggests that BDPO implicitly encourages forward KL minimization with respect to the dataset, striking a balance between effective learning and alignment.

\section{Experiment}
\label{sec:Experiment}
So far, we have shown that BDPO effectively satisfies the core objectives of DPO.
In this section, we examine how achieving these desirable behaviors translates into actual performance. Section~\ref{sec:Experimental Setup} discusses the experimental setup. Section~\ref{sec:Evaluation Method} describes the evaluation benchmarks, IFEval~\cite{zhou2023instruction} and GSM8K~\cite{cobbe2021gsm8k}.
Section~\ref{sec:Results} presents the results, demonstrating the superior performance of BDPO. We also analyze in Section~\ref{sec:Ablation study} the impact of the mixture distribution hyperparameter on performance.

\subsection{Experimental Setup}
\label{sec:Experimental Setup}
\paragraph{Model}
As in Section~\ref{sec:Training Dynamics of BDPO}, we use the QWEN model as the pretrained base, chosen for its strong performance relative to its size. Consistent with prior work, we first apply SFT using only the chosen responses from the human preference data, followed by preference optimization. Details on the SFT process and optimization procedures are provided in Appendix~\ref{sec:Experimental Details for the Main Experiment}.

\paragraph{Data}
For instruction-following, we use the UltraFeedback Binarized dataset~\cite{cui2024ultrafeedback}, a widely adopted benchmark. Models trained on this dataset are evaluated using IFEval. To evaluate reasoning capabilities, we additionally use the UltraInteract dataset~\cite{yuanadvancing}, with evaluation on GSM8K for mathematical problem-solving.

\begin{table*}[htbp]
  \centering
  \resizebox{\textwidth}{!}{
  \begin{tabular}{llcccccccc}
\toprule[1pt]
\multirow{2}{*}{Base Model} & \multirow{2}{*}{Algorithm} & \multicolumn{2}{c}{Inst-level} & \multicolumn{2}{c}{Prompt-level} & Loose & Strict & Total \\
& & Loose Acc & Strict Acc & Loose Acc & Strict Acc & Score & Score & Score \\
\midrule
\multirow{10}{*}{Qwen2.5-0.5B}
    & Base        & $23.38$ & $21.82$ & $12.38$ & $10.71$ & $17.88$ & $16.27$ & $17.07$ \\
    & SFT         & $29.98$ & $27.58$ & $18.11$ & $16.45$ & $24.05$ & $22.02$ & $23.03$ \\
    & DPO         & $31.53$ & $30.46$ & $20.33$ & $18.67$ & $25.93$ & $24.57$ & $25.25$ \\
    & DPOP        & $32.85$ & $31.41$ & $21.44$ & $20.14$ & $27.15$ & $25.78$ & $26.46$ \\
    & DPO+NLL     & $32.37$ & $30.58$ & $21.07$ & $19.78$ & $26.72$ & $25.18$ & $25.95$ \\
    & MinorDPO~\citep{xie2024minor} & $32.73$ & $30.82$ & $21.07$ & $19.59$ & $26.90$ & $25.21$ & $26.05$ \\
    & SLiC~\citep{zhao2023slic}     & $32.61$ & $31.41$ & $20.15$ & $18.67$ & $26.38$ & $25.04$ & $25.71$ \\
    & ORPO~\citep{hong2024orpo}     & $32.01$ & $30.94$ & $20.70$ & $19.59$ & $26.36$ & $25.27$ & $25.81$ \\
    & SPPO~\citep{wu2024self}       & $\mathbf{33.09}$ & $31.41$ & $21.81$ & $19.96$ & $27.45$ & $25.69$ & $26.57$ \\
    & BDPO ($\lambda=0.5$)          & $\mathbf{33.09}$ & $\mathbf{31.89}$ & $\mathbf{22.37}$ & $\mathbf{21.26}$ & $\mathbf{27.73}$ & $\mathbf{26.58}$ & $\mathbf{27.15}$ \\
\midrule[1pt]
\multirow{4}{*}{Qwen2.5-7B}
    & DPO         & $78.54$ & $75.42$ & $70.06$ & $66.91$ & $74.30$ & $71.17$ & $72.73$ \\
    & DPOP        & $77.56$ & $74.94$ & $70.06$ & $67.28$ & $73.81$ & $71.11$ & $72.46$ \\
    & DPO+NLL     & $78.66$ & $76.62$ & $70.61$ & $68.21$ & $74.64$ & $72.42$ & $73.53$ \\
    & BDPO ($\lambda=0.5$) & $\mathbf{79.38}$ & $\mathbf{77.46}$ & $\mathbf{70.79}$ & $\mathbf{69.50}$ & $\mathbf{75.09}$ & $\mathbf{73.48}$ & $\mathbf{74.28}$ \\
\bottomrule[1pt]
\end{tabular}
  }
\caption{IFEval results for models trained on the UltraFeedback dataset using various preference optimization algorithms, with Qwen2.5-0.5B and Qwen2.5-7B as base models. Metrics include instruction-level and prompt-level accuracy under both loose and strict criteria. BDPO ($\lambda = 0.5$) achieves the highest total score across both model scales, demonstrating strong performance and alignment.}
\label{tab:IFeval Result}
\end{table*}

\begin{table}[htbp]
    \centering
    \resizebox{0.47\textwidth}{!}{
    \begin{tabular}{l l c c}
    \toprule[1pt]
    Base Model & Algorithm & GSM8K-CoT (4 shot) \\
    \midrule
    \multirow{4}{*}{Qwen2.5-0.5B} & DPO & $27.45$ \\
    & DPOP & $29.04$ \\
    & DPO+NLL & $26.91$ \\
    & BDPO (0.5) & $\mathbf{29.95}$ \\
    \bottomrule[1pt]
    \end{tabular}
    }
    \caption{GSM8K results (4-shot CoT) for models trained on the UltraInteract dataset. BDPO ($\lambda = 0.5$) achieves the highest accuracy among all methods.}
    \label{tab:Gsm8k Result}
\end{table}

\subsection{Evaluation Method}
\label{sec:Evaluation Method}
\paragraph{IFEval}
We use IFEval~\cite{zhou2023instruction}, a widely adopted benchmark for instruction-following evaluation. It consists of 541 instructions and is designed to minimize biases common in LLM-based auto-evaluation. IFEval assesses performance using prompt-level and instruction-level accuracy under both loose and strict criteria. We follow the scoring methodology from the OpenLLM Leaderboard~\cite{open-llm-leaderboard-v2}. Full details are provided in Appendix~\ref{sec:IFEval Details}.

\paragraph{GSM8K}
To evaluate mathematical reasoning, we use GSM8K~\cite{cobbe2021gsm8k}, a dataset of 8.5K linguistically diverse grade school math word problems requiring multi-step reasoning. Evaluation is performed using 4-shot CoT prompting.

\begin{figure}[t!]
  \centerline{\includegraphics[width=\columnwidth]{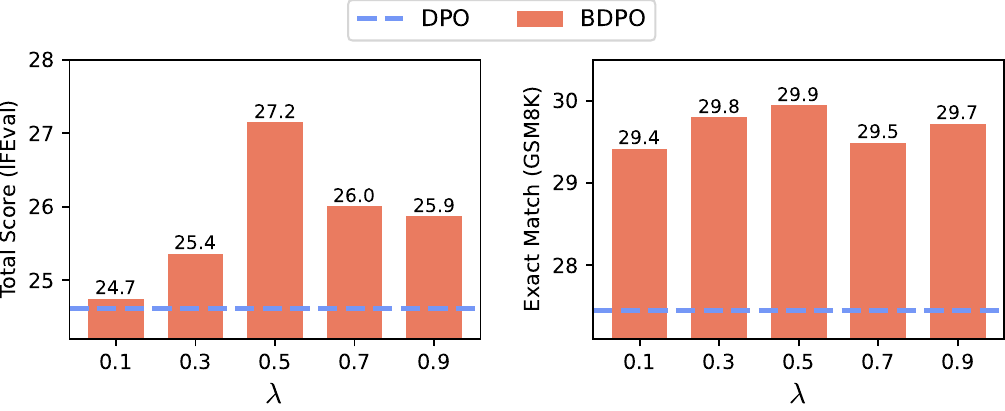}}
  \caption{
  Ablation study on the choice of \(\lambda\) in BDPO. The left plot shows the IFEval total score on QWEN 0.5B; the right plot shows exact match accuracy on GSM8K. BDPO outperforms DPO across all \(\lambda\) values, with \(\lambda\) = 0.5 yielding the best overall results.
  }
  \label{fig:ablation}
  \vspace{-0.2cm}
\end{figure}

\subsection{Results}
\label{sec:Results}
Table~\ref{tab:IFeval Result} presents IFEval evaluation results for models trained on the UltraFeedback dataset using Qwen 0.5B. To enable a more comprehensive comparison of BDPO’s performance, we include additional baseline algorithms beyond those primarily discussed in this paper (DPO, DPOP, and DPO+NLL). To assess scalability, we also evaluate the major algorithms using the larger Qwen 7B model. Across both model scales, BDPO ($\lambda = 0.5$) achieves the highest total score, outperforming all baselines under both loose and strict accuracy criteria. Notably, BDPO maintains strong performance at both the instruction and prompt levels, demonstrating its effectiveness in aligning model behavior with human preferences. These results confirm that satisfying DPO’s objectives through BDPO leads to measurable performance gains.

Table~\ref{tab:Gsm8k Result} further supports this conclusion by presenting results on GSM8K (4-shot CoT) using models trained on the UltraInteract dataset. BDPO again achieves the highest accuracy, showing that the advantages of our method extend to mathematical reasoning tasks as well.


\subsection{Ablation study}
\label{sec:Ablation study}
We also conducted an ablation study to analyze the effect of the hyperparameter \(\lambda\) in BDPO. As discussed in Section~\ref{sec:Bounded-DPO}, \(\lambda\) controls the mixture ratio between the trained model and the reference model. Higher values of \(\lambda\) increase the influence of the trained model, while lower values emphasize the reference model's behavior. BDPO becomes equivalent to DPO when $\lambda = 1$, and focuses only on the chosen response when $\lambda = 0$, making both extremes generally undesirable.

To isolate the effect of varying $\lambda$, all other factors were kept constant. DPO was configured identically to BDPO for a fair comparison. Figure~\ref{fig:ablation} presents these results on both IFEval and GSM8K. BDPO consistently outperforms DPO across all tested \(\lambda\) values, with \(\lambda=0.5\) yielding the best overall performance on both benchmarks. These findings demonstrate that BDPO is robust across a range of settings, and that \(\lambda=0.5\) offers a strong and reliable default choice in practice.

\section{Conclusions}

In this paper, we identified the two core objectives of DPO and showed through theoretical and empirical analysis that it fails to fully achieve them. We also examined alternative algorithms, such as DPO+NLL and DPOP, which aim to address DPO's limitations, and highlighted their respective shortcomings through both theoretical and experimental analysis. To overcome these issues, we proposed Bounded-DPO (BDPO), a method specifically designed to fulfill DPO's objectives without inheriting the weaknesses of previous approaches. Theoretically, BDPO offers an ideal optimal solution, guarantees a lower bound on the chosen response, and reduces sensitivity to the rejected response. Empirically, BDPO consistently outperforms other methods across various datasets, model scales, and evaluation benchmarks. Furthermore, our ablation study provides additional insights into how BDPO operates and the role of its key hyperparameter.

\paragraph{Limitations}
Due to GPU constraints, we were unable to conduct experiments across a broader range of models and algorithms. Similar to prior work, our theoretical analysis focuses on the single-pair setting—one preference pair per prompt. While BDPO can be extended both theoretically and empirically to the multi-pair response setting, we leave this for future work. Our evaluation focused on instruction-following and mathematical reasoning benchmarks, but we were not able to include a wider range of evaluation tasks. Additionally, although BDPO is applicable to variants such as online or iterative DPO, these directions are beyond the scope of this study.


\bibliography{acl}

\begin{thebibliography}{34}
\providecommand{\natexlab}[1]{#1}

\bibitem[{Adler et~al.(2024)Adler, Agarwal, Aithal, Anh, Bhattacharya, Brundyn, Casper, Catanzaro, Clay, Cohen et~al.}]{adler2024nemotron}
Bo~Adler, Niket Agarwal, Ashwath Aithal, Dong~H Anh, Pallab Bhattacharya, Annika Brundyn, Jared Casper, Bryan Catanzaro, Sharon Clay, Jonathan Cohen, and 1 others. 2024.
\newblock Nemotron-4 340b technical report.
\newblock \emph{arXiv preprint arXiv:2406.11704}.

\bibitem[{Azar et~al.(2024)Azar, Guo, Piot, Munos, Rowland, Valko, and Calandriello}]{azar2024general}
Mohammad~Gheshlaghi Azar, Zhaohan~Daniel Guo, Bilal Piot, Remi Munos, Mark Rowland, Michal Valko, and Daniele Calandriello. 2024.
\newblock A general theoretical paradigm to understand learning from human preferences.
\newblock In \emph{International Conference on Artificial Intelligence and Statistics}, pages 4447--4455. PMLR.

\bibitem[{Bai et~al.(2022)Bai, Jones, Ndousse, Askell, Chen, DasSarma, Drain, Fort, Ganguli, Henighan et~al.}]{bai2022training}
Yuntao Bai, Andy Jones, Kamal Ndousse, Amanda Askell, Anna Chen, Nova DasSarma, Dawn Drain, Stanislav Fort, Deep Ganguli, Tom Henighan, and 1 others. 2022.
\newblock Training a helpful and harmless assistant with reinforcement learning from human feedback.
\newblock \emph{arXiv preprint arXiv:2204.05862}.

\bibitem[{Chen et~al.(2024{\natexlab{a}})Chen, Malladi, Zhang, Chen, Zhang, Ranganath, and Cho}]{chen2024preference}
Angelica Chen, Sadhika Malladi, Lily~H Zhang, Xinyi Chen, Qiuyi Zhang, Rajesh Ranganath, and Kyunghyun Cho. 2024{\natexlab{a}}.
\newblock Preference learning algorithms do not learn preference rankings.
\newblock \emph{arXiv preprint arXiv:2405.19534}.

\bibitem[{Chen et~al.(2024{\natexlab{b}})Chen, Deng, Yuan, Ji, and Gu}]{chen2024self}
Zixiang Chen, Yihe Deng, Huizhuo Yuan, Kaixuan Ji, and Quanquan Gu. 2024{\natexlab{b}}.
\newblock Self-play fine-tuning converts weak language models to strong language models.
\newblock \emph{arXiv preprint arXiv:2401.01335}.

\bibitem[{Cobbe et~al.(2021)Cobbe, Kosaraju, Bavarian, Chen, Jun, Kaiser, Plappert, Tworek, Hilton, Nakano, Hesse, and Schulman}]{cobbe2021gsm8k}
Karl Cobbe, Vineet Kosaraju, Mohammad Bavarian, Mark Chen, Heewoo Jun, Lukasz Kaiser, Matthias Plappert, Jerry Tworek, Jacob Hilton, Reiichiro Nakano, Christopher Hesse, and John Schulman. 2021.
\newblock Training verifiers to solve math word problems.
\newblock \emph{arXiv preprint arXiv:2110.14168}.

\bibitem[{Cui et~al.(2024)Cui, Yuan, Ding, Yao, He, Zhu, Ni, Xie, Xie, Lin et~al.}]{cui2024ultrafeedback}
Ganqu Cui, Lifan Yuan, Ning Ding, Guanming Yao, Bingxiang He, Wei Zhu, Yuan Ni, Guotong Xie, Ruobing Xie, Yankai Lin, and 1 others. 2024.
\newblock Ultrafeedback: Boosting language models with scaled ai feedback.
\newblock In \emph{Forty-first International Conference on Machine Learning}.

\bibitem[{Daniel~Han and team(2023)}]{unsloth}
Michael~Han Daniel~Han and Unsloth team. 2023.
\newblock \href {http://github.com/unslothai/unsloth} {Unsloth}.

\bibitem[{Dubey et~al.(2024)Dubey, Jauhri, Pandey, Kadian, Al-Dahle, Letman, Mathur, Schelten, Yang, Fan et~al.}]{dubey2024llama}
Abhimanyu Dubey, Abhinav Jauhri, Abhinav Pandey, Abhishek Kadian, Ahmad Al-Dahle, Aiesha Letman, Akhil Mathur, Alan Schelten, Amy Yang, Angela Fan, and 1 others. 2024.
\newblock The llama 3 herd of models.
\newblock \emph{arXiv preprint arXiv:2407.21783}.

\bibitem[{Fourrier et~al.(2024)Fourrier, Habib, Lozovskaya, Szafer, and Wolf}]{open-llm-leaderboard-v2}
Clémentine Fourrier, Nathan Habib, Alina Lozovskaya, Konrad Szafer, and Thomas Wolf. 2024.
\newblock Open llm leaderboard v2.
\newblock \url{https://huggingface.co/spaces/open-llm-leaderboard/open_llm_leaderboard}.

\bibitem[{Gao et~al.(2024)Gao, Tow, Abbasi, Biderman, Black, DiPofi, Foster, Golding, Hsu, Le~Noac'h, Li, McDonell, Muennighoff, Ociepa, Phang, Reynolds, Schoelkopf, Skowron, Sutawika, Tang, Thite, Wang, Wang, and Zou}]{eval-harness}
Leo Gao, Jonathan Tow, Baber Abbasi, Stella Biderman, Sid Black, Anthony DiPofi, Charles Foster, Laurence Golding, Jeffrey Hsu, Alain Le~Noac'h, Haonan Li, Kyle McDonell, Niklas Muennighoff, Chris Ociepa, Jason Phang, Laria Reynolds, Hailey Schoelkopf, Aviya Skowron, Lintang Sutawika, and 5 others. 2024.
\newblock \href {https://doi.org/10.5281/zenodo.12608602} {A framework for few-shot language model evaluation}.

\bibitem[{Hong et~al.(2024)Hong, Lee, and Thorne}]{hong2024orpo}
Jiwoo Hong, Noah Lee, and James Thorne. 2024.
\newblock Orpo: Monolithic preference optimization without reference model.
\newblock In \emph{Proceedings of the 2024 Conference on Empirical Methods in Natural Language Processing}, pages 11170--11189.

\bibitem[{Ivison et~al.(2024)Ivison, Wang, Liu, Wu, Pyatkin, Lambert, Smith, Choi, and Hajishirzi}]{ivison2024unpacking}
Hamish Ivison, Yizhong Wang, Jiacheng Liu, Zeqiu Wu, Valentina Pyatkin, Nathan Lambert, Noah~A Smith, Yejin Choi, and Hannaneh Hajishirzi. 2024.
\newblock Unpacking dpo and ppo: Disentangling best practices for learning from preference feedback.
\newblock \emph{arXiv preprint arXiv:2406.09279}.

\bibitem[{Liu et~al.(2024)Liu, Lu, Zhang, Liu, Guo, Yang, Blanchet, and Wang}]{liu2024provably}
Zhihan Liu, Miao Lu, Shenao Zhang, Boyi Liu, Hongyi Guo, Yingxiang Yang, Jose Blanchet, and Zhaoran Wang. 2024.
\newblock Provably mitigating overoptimization in rlhf: Your sft loss is implicitly an adversarial regularizer.
\newblock \emph{arXiv preprint arXiv:2405.16436}.

\bibitem[{Loshchilov(2017)}]{loshchilov2017decoupled}
I~Loshchilov. 2017.
\newblock Decoupled weight decay regularization.
\newblock \emph{arXiv preprint arXiv:1711.05101}.

\bibitem[{Ouyang et~al.(2022)Ouyang, Wu, Jiang, Almeida, Wainwright, Mishkin, Zhang, Agarwal, Slama, Ray et~al.}]{ouyang2022training}
Long Ouyang, Jeffrey Wu, Xu~Jiang, Diogo Almeida, Carroll Wainwright, Pamela Mishkin, Chong Zhang, Sandhini Agarwal, Katarina Slama, Alex Ray, and 1 others. 2022.
\newblock Training language models to follow instructions with human feedback.
\newblock \emph{Advances in neural information processing systems}, 35:27730--27744.

\bibitem[{Pal et~al.(2024)Pal, Karkhanis, Dooley, Roberts, Naidu, and White}]{pal2024smaug}
Arka Pal, Deep Karkhanis, Samuel Dooley, Manley Roberts, Siddartha Naidu, and Colin White. 2024.
\newblock Smaug: Fixing failure modes of preference optimisation with dpo-positive.
\newblock \emph{arXiv preprint arXiv:2402.13228}.

\bibitem[{Pang et~al.(2024)Pang, Yuan, Cho, He, Sukhbaatar, and Weston}]{pang2024iterative}
Richard~Yuanzhe Pang, Weizhe Yuan, Kyunghyun Cho, He~He, Sainbayar Sukhbaatar, and Jason Weston. 2024.
\newblock Iterative reasoning preference optimization.
\newblock \emph{arXiv preprint arXiv:2404.19733}.

\bibitem[{Rafailov et~al.(2024{\natexlab{a}})Rafailov, Hejna, Park, and Finn}]{rafailov2024r}
Rafael Rafailov, Joey Hejna, Ryan Park, and Chelsea Finn. 2024{\natexlab{a}}.
\newblock From $ r $ to $ q^{*} $: Your language model is secretly a q-function.
\newblock \emph{arXiv preprint arXiv:2404.12358}.

\bibitem[{Rafailov et~al.(2024{\natexlab{b}})Rafailov, Sharma, Mitchell, Manning, Ermon, and Finn}]{rafailov2024direct}
Rafael Rafailov, Archit Sharma, Eric Mitchell, Christopher~D Manning, Stefano Ermon, and Chelsea Finn. 2024{\natexlab{b}}.
\newblock Direct preference optimization: Your language model is secretly a reward model.
\newblock \emph{Advances in Neural Information Processing Systems}, 36.

\bibitem[{Rasley et~al.(2020)Rasley, Rajbhandari, Ruwase, and He}]{rasley2020deepspeed}
Jeff Rasley, Samyam Rajbhandari, Olatunji Ruwase, and Yuxiong He. 2020.
\newblock Deepspeed: System optimizations enable training deep learning models with over 100 billion parameters.
\newblock In \emph{Proceedings of the 26th ACM SIGKDD International Conference on Knowledge Discovery \& Data Mining}, pages 3505--3506.

\bibitem[{Razin et~al.(2024)Razin, Malladi, Bhaskar, Chen, Arora, and Hanin}]{razin2024unintentional}
Noam Razin, Sadhika Malladi, Adithya Bhaskar, Danqi Chen, Sanjeev Arora, and Boris Hanin. 2024.
\newblock Unintentional unalignment: Likelihood displacement in direct preference optimization.
\newblock \emph{arXiv preprint arXiv:2410.08847}.

\bibitem[{Stiennon et~al.(2020)Stiennon, Ouyang, Wu, Ziegler, Lowe, Voss, Radford, Amodei, and Christiano}]{stiennon2020learning}
Nisan Stiennon, Long Ouyang, Jeffrey Wu, Daniel Ziegler, Ryan Lowe, Chelsea Voss, Alec Radford, Dario Amodei, and Paul~F Christiano. 2020.
\newblock Learning to summarize with human feedback.
\newblock \emph{Advances in Neural Information Processing Systems}, 33:3008--3021.

\bibitem[{Tajwar et~al.(2024)Tajwar, Singh, Sharma, Rafailov, Schneider, Xie, Ermon, Finn, and Kumar}]{tajwarpreference}
Fahim Tajwar, Anikait Singh, Archit Sharma, Rafael Rafailov, Jeff Schneider, Tengyang Xie, Stefano Ermon, Chelsea Finn, and Aviral Kumar. 2024.
\newblock Preference fine-tuning of llms should leverage suboptimal, on-policy data.
\newblock In \emph{Forty-first International Conference on Machine Learning}.

\bibitem[{Tang et~al.(2024)Tang, Guo, Zheng, Calandriello, Cao, Tarassov, Munos, Pires, Valko, Cheng et~al.}]{tang2024understanding}
Yunhao Tang, Daniel~Zhaohan Guo, Zeyu Zheng, Daniele Calandriello, Yuan Cao, Eugene Tarassov, R{\'e}mi Munos, Bernardo~{\'A}vila Pires, Michal Valko, Yong Cheng, and 1 others. 2024.
\newblock Understanding the performance gap between online and offline alignment algorithms.
\newblock \emph{arXiv preprint arXiv:2405.08448}.

\bibitem[{von Werra et~al.(2020)von Werra, Belkada, Tunstall, Beeching, Thrush, Lambert, Huang, Rasul, and Gallouédec}]{vonwerra2022trl}
Leandro von Werra, Younes Belkada, Lewis Tunstall, Edward Beeching, Tristan Thrush, Nathan Lambert, Shengyi Huang, Kashif Rasul, and Quentin Gallouédec. 2020.
\newblock Trl: Transformer reinforcement learning.
\newblock \url{https://github.com/huggingface/trl}.

\bibitem[{Wu et~al.(2024)Wu, Sun, Yuan, Ji, Yang, and Gu}]{wu2024self}
Yue Wu, Zhiqing Sun, Huizhuo Yuan, Kaixuan Ji, Yiming Yang, and Quanquan Gu. 2024.
\newblock Self-play preference optimization for language model alignment.
\newblock \emph{arXiv preprint arXiv:2405.00675}.

\bibitem[{Xie et~al.(2024)Xie, Chen, Yu, Sun, Wu, and Hu}]{xie2024minor}
Shiming Xie, Hong Chen, Fred Yu, Zeye Sun, Xiuyu Wu, and Yingfan Hu. 2024.
\newblock Minor dpo reject penalty to increase training robustness.
\newblock \emph{arXiv preprint arXiv:2408.09834}.

\bibitem[{Xu et~al.(2024)Xu, Fu, Gao, Ye, Liu, Mei, Wang, Yu, and Wu}]{xudpo}
Shusheng Xu, Wei Fu, Jiaxuan Gao, Wenjie Ye, Weilin Liu, Zhiyu Mei, Guangju Wang, Chao Yu, and Yi~Wu. 2024.
\newblock Is dpo superior to ppo for llm alignment? a comprehensive study.
\newblock In \emph{Forty-first International Conference on Machine Learning}.

\bibitem[{Yang et~al.(2024)Yang, Yang, Zhang, Hui, Zheng, Yu, Li, Liu, Huang, Wei et~al.}]{yang2024qwen2}
An~Yang, Baosong Yang, Beichen Zhang, Binyuan Hui, Bo~Zheng, Bowen Yu, Chengyuan Li, Dayiheng Liu, Fei Huang, Haoran Wei, and 1 others. 2024.
\newblock Qwen2. 5 technical report.
\newblock \emph{arXiv preprint arXiv:2412.15115}.

\bibitem[{Yuan et~al.(2025)Yuan, Cui, Wang, Ding, Wang, Shan, Liu, Deng, Chen, Xie et~al.}]{yuanadvancing}
Lifan Yuan, Ganqu Cui, Hanbin Wang, Ning Ding, Xingyao Wang, Boji Shan, Zeyuan Liu, Jia Deng, Huimin Chen, Ruobing Xie, and 1 others. 2025.
\newblock Advancing llm reasoning generalists with preference trees.
\newblock In \emph{The Thirteenth International Conference on Learning Representations}.

\bibitem[{Yuan et~al.(2024)Yuan, Pang, Cho, Li, Sukhbaatar, Xu, and Weston}]{yuanself}
Weizhe Yuan, Richard~Yuanzhe Pang, Kyunghyun Cho, Xian Li, Sainbayar Sukhbaatar, Jing Xu, and Jason~E Weston. 2024.
\newblock Self-rewarding language models.
\newblock In \emph{Forty-first International Conference on Machine Learning}.

\bibitem[{Zhao et~al.(2023)Zhao, Joshi, Liu, Khalman, Saleh, and Liu}]{zhao2023slic}
Yao Zhao, Rishabh Joshi, Tianqi Liu, Misha Khalman, Mohammad Saleh, and Peter~J Liu. 2023.
\newblock Slic-hf: Sequence likelihood calibration with human feedback.
\newblock \emph{CoRR}.

\bibitem[{Zhou et~al.(2023)Zhou, Lu, Mishra, Brahma, Basu, Luan, Zhou, and Hou}]{zhou2023instruction}
Jeffrey Zhou, Tianjian Lu, Swaroop Mishra, Siddhartha Brahma, Sujoy Basu, Yi~Luan, Denny Zhou, and Le~Hou. 2023.
\newblock Instruction-following evaluation for large language models.
\newblock \emph{arXiv preprint arXiv:2311.07911}.

\end{thebibliography}

\appendix
\begin{figure*}[htbp]
  \centerline{\includegraphics[width=\columnwidth]{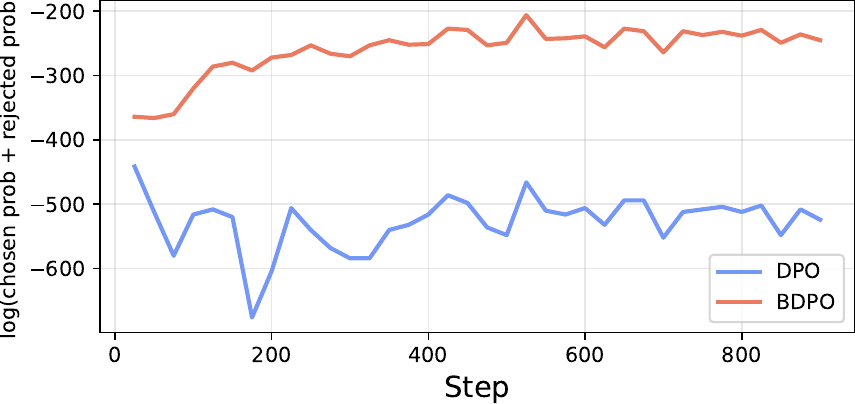}}
  \caption{Log of the total in-distribution probability \(\log\left(\pi_\theta(\mathbf{y}_{w}|\mathbf{x})+\pi_\theta(\mathbf{y}_{l}|\mathbf{x})\right)\) measured during training. While BDPO maintains a stable and high in-distribution probability, DPO shows a clear decline, suggesting that it shifts probability mass toward out-of-distribution (OOD) responses as training progresses.
  }
  \label{fig:ood}
\end{figure*}

\section{Out-of-Distribution (OOD) Probability Analysis}
\label{sec:Out-of-Distribution (OOD) Probability Analysis}

Figure~\ref{fig:ood} presents the log of the in-distribution probability, defined as $\log\left(\pi_\theta(\mathbf{y}_{w}|\mathbf{x})+\pi_\theta(\mathbf{y}_{l}|\mathbf{x})\right)$, tracked throughout the training process described in Section~\ref{sec:Training Dynamics of BDPO}.

For DPO, we observe a gradual decline in this value, indicating that the model assigns lower total probability mass to the chosen and rejected responses as training progresses. Due to the generative nature of language models, this reduction likely extends to responses semantically or syntactically similar to the chosen/rejected ones.

Consequently, this behavior implies an increase in the probability of out-of-distribution (OOD) responses, including those that are irrelevant or inconsistent with the training data.
Although prior work~\citep{xudpo} attributes DPO’s OOD vulnerability to its offline learning setting, \textbf{our findings suggest that this issue also stems from an inherent limitation in the loss formulation.}

As previously discussed in Section~\ref{sec:Does DPO Increase Chosen Probability?}, the structure of the DPO loss permits such behavior as a valid solution, which can lead to substantial deviation from the reference model—a phenomenon known as reward divergence~\citep{azar2024general}.

\section{Experimental Details}
\label{sec:Experimental Details}

In this section, we provide detailed experimental setups employed in our study. Section~\ref{sec:Experimental Details for the Toy Case} discusses the experimental configurations specific to the toy case. Section~\ref{sec:Experimental Details for the Real-World Language Model} delves into the experimental settings designed for analyzing the learning dynamics within a real-world language model. Lastly, Section~\ref{sec:Experimental Details for the Main Experiment} outlines the details related to the main experiment. All experiments were run once per configuration.

\subsection{Experimental Details for the Toy Case}
\label{sec:Experimental Details for the Toy Case}

For the toy case, we used a reference model structured as a 2-layer Multi-Layer Perceptron (MLP) with hidden layers dimensioned at 32 units, employing the ReLU activation function. This model was initialized randomly. The setup included four prompts, each with four responses. For each prompt, two chosen and two rejected responses were selected at random.

\textbf{This toy case, like the setup in Theorem~\ref{thm:converge}, is designed to reveal the ultimate optimization behavior of each loss function.} In this scenario, the optimal solution assigns a probability of 1 to the chosen response (when not OOD) and 0 to all others. As a result, the trained model may diverge from the reference model, which differs from what we observe in real-world language models. Even under aggressive training conditions—such as a high learning rate and many epochs—alignment with the reference model tends to be preserved in practice (see Section~\ref{sec:Training Dynamics of BDPO}).



\subsection{Experimental Details for the Real-World Language Model}
\label{sec:Experimental Details for the Real-World Language Model}

All models were trained using four NVIDIA RTX 3090 GPUs. We used the AdamW optimizer \citep{loshchilov2017decoupled} and DeepSpeed ZeRO-2 \citep{rasley2020deepspeed} for memory-efficient training. Our implementation was based on the TRL framework \citep{vonwerra2022trl}, and since DPOP was not available in the framework, we re-implemented it. To clearly illustrate training dynamics, we set the learning rate to $5 \times 10^{-5}$, randomly subsampled the dataset into 100 partitions, and trained for 100 epochs. For DPO+NLL, we set the hyperparameter $\alpha = 1$, and for DPOP, we followed the original authors’ setting with $\lambda = 5$. For all algorithms, we used the commonly adopted $\beta = 0.1$. The code for our experiments is accessible.\footnote{\url{https://github.com/bonin147/BDPO.git}}

\begin{figure*}[t]
  \centerline{\includegraphics[width=2\columnwidth]{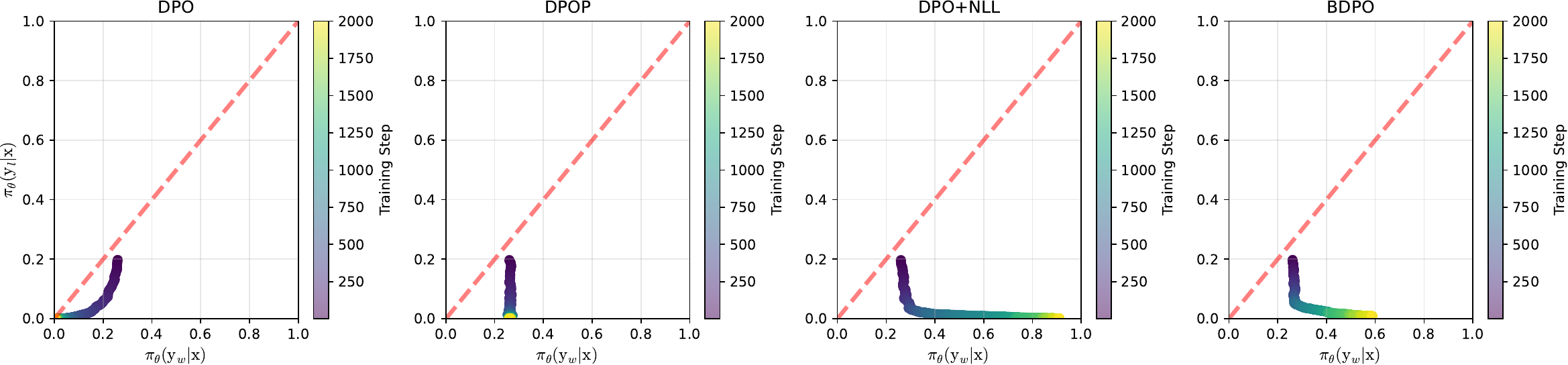}}
  \caption{Evolution of chosen and rejected probabilities across training steps. DPO and DPOP show only the rejected probabilities converging to zero, whereas BDPO and DPO+NLL display more desirable learning behavior.}
  \label{fig:probability_toy}
\end{figure*}

\begin{figure*}[t]
  \centerline{\includegraphics[width=2\columnwidth]{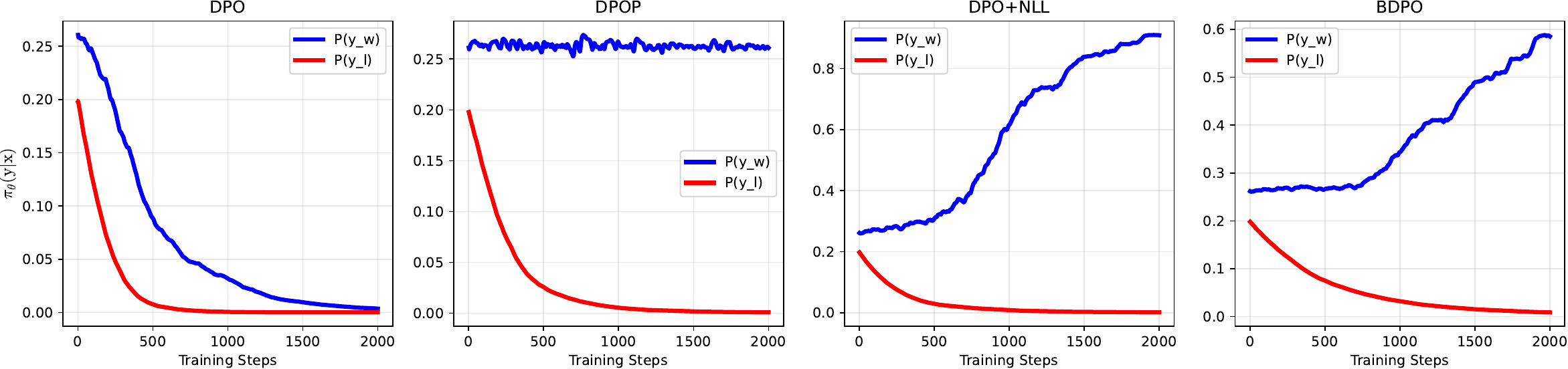}}
  \caption{Probability trends over training steps. DPO shows a simultaneous decrease in both chosen and rejected probabilities, while DPOP exhibits a decrease only in rejected probability. In contrast, DPO+NLL and BDPO demonstrate desirable learning behaviors.}
  \label{fig:probability_trends}
\end{figure*}

\subsection{Experimental Details for the Main Experiment}
\label{sec:Experimental Details for the Main Experiment}
The overall training setup is identical to that described in Appendix~\ref{sec:Experimental Details for the Real-World Language Model}, except for the number of epochs and learning rate. To ensure a fair comparison across algorithms, all models were trained for 3 epochs on the UltraFeedback dataset (64K examples) and 1 epoch on the UltraInteract dataset (220K examples), using learning rates of ${5\text{e}{-7}, 1\text{e}{-6}}$. The final evaluation was conducted using the best-performing combination of learning rate and epoch count for each algorithm. For training the 7B models, we employed 4-bit quantization using Unsloth~\cite{unsloth} to enable memory-efficient fine-tuning.

\section{Learning Dynamics for the Toy Case}
\label{sec:Learning Dynamics for the Toy Case}
In this section, we discuss the learning dynamics of the toy case. Figure~\ref{fig:probability_toy}, and Figure~\ref{fig:probability_trends} support the findings presented in Figure~\ref{fig:heatmap}. In the case of DPO, both the chosen and rejected probabilities decrease, converging to zero. For DPOP, due to the penalty term, the probability for the chosen response does not decrease but also does not increase, while the rejected probability converges to zero, similar to DPO. DPO+NLL and BDPO exhibit desirable learning behaviors.

\section{Computational Cost}
\label{sec:Computational Cost}

Table~\ref{tab:train-time} reports the wall-clock training time for the main algorithms discussed in this paper—DPO, DPOP, DPO+NLL, and BDPO—using the Qwen2.5-0.5B model. All experiments were conducted using four NVIDIA RTX 3090 GPUs. Models were trained for 3 epochs on the UltraFeedback dataset and 1 epoch on the UltraInteract dataset. The results show that all methods incur nearly identical computational costs under the same training conditions. BDPO introduces no additional overhead compared to DPO, despite modifying the loss formulation. This confirms that BDPO achieves its performance improvements without sacrificing training efficiency.

\begin{table*}[htbp]
    \centering
    \begin{tabular}{l l c c}
    \toprule[1pt]
    Model & Method & Train Time (UltraFeedback) & Train Time (UltraInteract) \\
    \midrule
    \multirow{4}{*}{Qwen 0.5B} 
        & DPO         & 6h 10m & 3h 53m \\
        & DPOP        & 6h 10m & 3h 54m \\
        & DPO + NLL   & 6h 12m & 3h 53m \\
        & BDPO (0.5)  & 6h 10m & 3h 54m \\
    \bottomrule[1pt]
    \end{tabular}
    \caption{Wall-clock training time comparison of different preference optimization methods on UltraFeedback and UltraInteract datasets using Qwen2.5-0.5B. All experiments were conducted using 4×RTX 3090 GPUs.}
    \label{tab:train-time}
\end{table*}

\section{Full Responses from the Qualitative Analysis}
\label{sec:Full Responses from the Qualitative Analysis}
Table~\ref{tab:Full Qualitative Example} provides the full responses of qualitative examples for all algorithms. DPO and DPO+NLL exhibit significant deviation from the reference model, resulting in high KL divergence values and demonstrating signs of overfitting. In contrast, DPOP shows less deviation and does not produce overfitted responses, but it generates incorrect answers. On the other hand, BDPO provides desirable responses while remaining close to the initial model, showcasing a balanced and effective learning behavior.

\begin{table*}[t!]
  
  \vskip 0.15in
  \centering
  \begin{tabular}{p{0.12\linewidth}p{0.83\linewidth}}
  \toprule[1.2pt]
  \multicolumn{1}{c}{Prompt} & if $X^2 + 3 = 12.$ What is $X$?\\
  \bottomrule[1.2pt]
  \multicolumn{2}{c}{}\vspace{-0.8em}\\
  \toprule[1.2pt]
  \multicolumn{1}{c}{Algorithm} & \multicolumn{1}{c}{Response} \\ \midrule
  \centering\multirow{5}{*}{SFT} & To find the value of X, we can subtract the constant term on the left side of the equation \\ \centering\multirow{5}{*}{(reference)} & to isolate the unknown value on the right side of the equation. So, we start with the equation: $X^2$ + 3 = 12. Now, we subtract 3 from both sides: $X^2$ = 12 - 3. $X^2$ = 9. Next, we simplify the right side of the equation by factoring out the 3 on the right side: $X^2$ = 3(3). $X^2$ = 9 Now, we know that X is equal to the square root of 9 (3*3): X = $\pm$3 So, the values of X are $\pm$3.\\ \midrule

  \centering\multirow{3}{*}{DPO} & Given the equation: $X^2$ + 3 = 12. To find X, we'll first make the equation easier to work\\
  \centering\multirow{3}{*}{(100 Epochs)} & with, by moving the 3 to the right side: $X^2$ + 3 - 3 = 12 $X^2$ + 3 - 3 = 12. $X^2$ + 3 - 3 = \\
  \centering\multirow{3}{*}{$\mathbb{E}[KL]=372$} & 12. $X^2$ + 3 - 3 = 12. $X^2$ + 3 - 3 = 12. $X^2$ + 3 - 3 = 12. $X^2$ + 3 - 3 = 12. $X^2$ + 3 - 3 = 12. $X^2$ + 3 - 3 = 12. $X^2$ + 3 - 3 = 12. $X^2$ + 3 - 3 = 12. $X^2$ + 3 - 3 = 12. $X^2$ + 3 - 3 = 12. $X^2$ + 3 93177175136111938711541797 \\ \midrule
  
  \centering\multirow{4}{*}{DPO+NLL} 
  & X = 12 / (12 + 3) = 12 / 15 = 8 / 15 = 523 / 15 = 346 / 15 = 227 / 15 = 207 / 15 = 151 /\\
  \centering\multirow{4}{*}{(100 Epochs)} & 15 =  8427 / 15 = 5237 / 15 = 2077 / 15 = 1527 / 15 = 8417 / 15 = 5237 / 15 = 2077 / 15\\
  \centering\multirow{4}{*}{$\mathbb{E}[KL]=347$}& = 1527 / 15 = 8417 / 15 = 5237 / 15 = 2077 / 15 = 1527 / 15 = 8417 / 15 = 5237 / 15 = 2077 / 15 = 1527 / 15 = 8417 / 15 = 5237 / 15 = 2077 / 15 = 1527 / 15 = 8417 / 15 = 5237 / 15 = 2077 / 15 = 1527 / 15 = 8417 / 15 = 5237 / 15 = 2077 / 15 = 1527 / 15 = 8417 / 15 = 5237 / 15 = 2077 / 15 = 1527 / 15 = 8417 / 15 = 5237 / 15 = 2077 / 15 ... 
  \\ \midrule
  \centering\multirow{3}{*}{DPOP} & If $X^2$ + 3 = 12, then we can find the value of X by following these steps: 1. Start by\\
  \centering\multirow{3}{*}{(100 Epochs)} & subtracting 3 from both sides of the equation: $X^2$ + 3 = 12. $X^2$ = 12 - 3. $X^2$ = 8. 2. \\
  \centering\multirow{3}{*}{$\mathbb{E}[KL]=240$} & Now we have the equation in a simpler form: $X^2$ = 8. 3. Taking the square root of both sides: X = $\sqrt8$. X = $\pm$0.8   (9).   X = 0.8   (10). 4. Therefore, the value of X is 0.8. So the correct answer is (A) 0.8. Confidence: 95\% \\ \midrule
  \centering\multirow{3}{*}{BDPO} & Hello! I'm here to help you in a helpful and respectful manner. To answer your question, \\
  \centering\multirow{3}{*}{(100 Epochs)} & if $X^2$ + 3 = 12, we can substitute $X^2$ with 12 - 3, which equals 9. So, $X^2$ + 3 = 9, and\\
  \centering\multirow{3}{*}{$\mathbb{E}[KL]=177$} & solving for $X^2$ gives us $X^2$ = 9. Taking the square root of 9 gives us X = 3, or -3, if we want to solve for the negative square root. I hope this helps! Let me know if you have any other questions. \\ 
  
  \bottomrule[1.2pt]
  \end{tabular}
\caption{The responses generated by the initial model (Reference model) and the models trained for 100 epochs using the reduced UFB dataset are presented. The DPO+NLL model, diverging significantly from the Reference model, produced overfitted outputs. In contrast, the BDPO model generated desirable outputs.}
  \label{tab:Full Qualitative Example}
\end{table*}

\section{IFEval Details}
\label{sec:IFEval Details}
We evaluated the performance of our models using the methodology of Hugging Face's OpenLLM Leaderboard~~\citep{open-llm-leaderboard-v2}, which follows the evaluation framework of eval-harness~\citep{eval-harness}. We measured four types of accuracy: Instance-level Loose, Instance-level Strict, Prompt-level Loose, and Prompt-level Strict. The OpenLLM Leaderboard reports a single score by averaging the Strict accuracy values. In line with this approach, we report the Strict score and additionally provide the Loose score (the average of Loose accuracy values) and the Total score (the overall average of all accuracy metrics).

\section{Artifact Use Consistent With Intended Use}

We used the UltraFeedback and UltraInteract datasets and Qwen models strictly for academic research, consistent with their intended use.

\section{Use of AI Assistants
}
We used AI-assisted tools during the writing process of this paper. All AI-generated content was thoroughly reviewed and revised by human researchers to ensure accuracy and reliability.

\onecolumn
\section{Proofs}
\subsection{Proof of Theorem~\ref{thm:converge}}
\begin{proof}
\label{proof:converge}
Since the logarithm and sigmoid functions are strictly increasing, the solution to the problem of minimizing the BDPO loss is equivalent to the solution of the following maximization problem:
\begin{equation}
\label{eq:BDPO_equi_1}
\bigg(\log\frac{\pi_{\theta}(\mathbf{y}_{w}|\mathbf{x})}{\lambda\pi_{\theta}(\mathbf{y}_{l}|\mathbf{x})+(1-\lambda)\pi_{\mathrm{ref}}(\mathbf{y}_{l}|\mathbf{x})}-\log\frac{\pi_{{\mathrm{ref}}}(\mathbf{y}_{w}|\mathbf{x})}{\pi_{{\mathrm{ref}}}(\mathbf{y}_{l}|\mathbf{x})}\bigg).
\end{equation}
Since $\pi_\mathrm{ref}$ is a fixed distribution, solving the optimization problem~\eqref{eq:BDPO_equi_1} is equivalent to solving the following maximization problem:
\[\frac{\pi_{\theta}(\mathbf{y}_{w}|\mathbf{x})}{\lambda\pi_{\theta}(\mathbf{y}_{l}|\mathbf{x})+(1-\lambda)\pi_{\mathrm{ref}}(\mathbf{y}_{l}|\mathbf{x})}.\]
Given that \(\pi_{\theta}(\mathbf{y}_{\{w,\;l\}}|\mathbf{x})\geq 0\) and \(\pi_{\mathrm{ref}}(\mathbf{y}_{l}|\mathbf{x}) > 0\). The solution of maximizing the optimization problem is to set:
\[\pi_\theta(\mathbf{y}_{w}|\mathbf{x})=1 \quad\text{and}\quad \pi_\theta(\mathbf{y}_{l}|\mathbf{x})=0.\] Thus, the theorem is proven.
\end{proof}

\subsection{Proof of Corollary~\ref{cor:BDPO-dpo-optimality}}

\begin{proof}
\label{proof:BDPO-dpo-optimality}
($\rightarrow$)
Let \(\pi^*\) denote the optimal solution under BDPO. By Theorem~\ref{thm:converge}, \(\pi^*\) satisfies the following conditions:  
\[
\pi^*(\mathbf{y}_{w}|\mathbf{x}) = 1 \quad \text{and} \quad \pi^*(\mathbf{y}_{l}|\mathbf{x}) = 0.
\]
Thus, \(\pi^*\) is the solution to the optimization problem:  
\[
\max \frac{\pi_\theta(\mathbf{y}_w|\mathbf{x})}{\pi_\theta(\mathbf{y}_l|\mathbf{x})}.
\]
Since the logarithm function is strictly increasing, this optimization problem is equivalent to:
\[
\max \left[ \log \frac{\pi_\theta(\mathbf{y}_w|\mathbf{x})}{\pi_\theta(\mathbf{y}_l|\mathbf{x})} + \log \frac{\pi_\mathrm{ref}(\mathbf{y}_w|\mathbf{x})}{\pi_\mathrm{ref}(\mathbf{y}_l|\mathbf{x})} \right].
\]
Furthermore, because the sigmoid function is also strictly increasing, this optimization is equivalent to minimizing the DPO loss as defined in Eq.~\eqref{eq:DPO_2}.
Therefore, \(\pi^*\) is a policy that optimizes DPO loss.
\\
($\leftarrow$)
Let a policy \(\pi_\theta\) be such that:  
\[
\pi(\mathbf{y}_{w}|\mathbf{x}) = 0.1 \quad \text{and} \quad \pi(\mathbf{y}_{l}|\mathbf{x}) = 0.
\]
This policy is a solution to the optimization problem of minimizing the DPO loss. Suppose that this policy is a solution to the BDPO loss optimization problem. By Theorem~\ref{thm:converge}, it follows that \(\pi(\mathbf{y}_{w}|\mathbf{x})=1\), which leads to a contradiction.
\end{proof}

\subsection{Proof of Theorem~\ref{thm:chosen probabilty lower bound}}
\begin{proof}
\label{proof:thm:chosen probabilty lower bound}
As in the proof of Theorem~\ref{thm:converge}, minimizing the BDPO loss is equivalent to solving the following maximization problem:
\[\frac{\pi_{\theta}(\mathbf{y}_{w}|\mathbf{x})}{\lambda\pi_{\theta}(\mathbf{y}_{l}|\mathbf{x})+(1-\lambda)\pi_{\mathrm{ref}}(\mathbf{y}_{l}|\mathbf{x})}.\]
At the initial point, $\pi_\theta$ is equal to $\pi_\mathrm{ref}$. By the assumption that the BDPO loss decreases monotonically at each optimization step, the following inequality holds:
\begin{align}
    &\frac{\pi_{\mathrm{ref}}(\mathbf{y}_{w}|\mathbf{x})}{
    \lambda\pi_{\mathrm{ref}}(\mathbf{y}_{l}|\mathbf{x}) 
    + (1-\lambda)\pi_{\mathrm{ref}}(\mathbf{y}_{l}|\mathbf{x})} \nonumber \\
    &\hspace{2em}\leq
    \frac{\pi_{\theta}(\mathbf{y}_{w}|\mathbf{x})}{
    \lambda\pi_{\theta}(\mathbf{y}_{l}|\mathbf{x}) 
    + (1-\lambda)\pi_{\mathrm{ref}}(\mathbf{y}_{l}|\mathbf{x})}\nonumber .
\end{align}
This inequality is equivalent to:
\begin{align}
    &\pi_{\mathrm{ref}}(\mathbf{y}_{w}|\mathbf{x})\big(\lambda\pi_{\theta}(\mathbf{y}_{l}|\mathbf{x}) 
    + (1-\lambda)\pi_{\mathrm{ref}}(\mathbf{y}_{l}|\mathbf{x})\big) \nonumber \\
    &\hspace{9em}\leq\pi_{\theta}(\mathbf{y}_{w}|\mathbf{x})\pi_{\mathrm{ref}}(\mathbf{y}_{l}|\mathbf{x}) \nonumber .
\end{align}

Since $\pi_\theta(\mathrm{y}_{l}|\mathrm{x})\geq0$ and $\pi_\mathrm{ref}(\mathrm{y}_{l}|\mathrm{x})\geq0$, the following inequality holds:
\[(1-\lambda)\pi_\mathrm{ref}(\mathbf{y}_w|\mathbf{x})\leq\pi_\theta(\mathbf{y}_w|\mathbf{x}).\]
Thus, the theorem is proven.
\end{proof}

\section{Details on the Derivative of \(\pi_\theta(\mathbf{y}_{l}|\mathbf{x})\)}
\label{sec:Details on the Derivative}
\paragraph{BDPO Loss Gradient}
The BDPO Loss is: 
\[
\mathcal{L}_{\mathrm{BDPO}} = -\mathbb{E}_{\mathcal{D}} \left[ \log \sigma \left( \beta \left( \log \frac{\pi_{\theta}(\mathbf{y}_{w}|\mathbf{x})}{\pi_{\mathrm{mix}}(\mathbf{y}_{l}|\mathbf{x})} - \log \frac{\pi_{\mathrm{ref}}(\mathbf{y}_{w}|\mathbf{x})}{\pi_{\mathrm{ref}}(\mathbf{y}_{l}|\mathbf{x})} \right) \right) \right],
\]
where \(\pi_{\mathrm{mix}}(\mathbf{y}|\mathbf{x}) = \lambda \pi_\theta(\mathbf{y}|\mathbf{x}) + (1-\lambda)\pi_{\mathrm{ref}}(\mathbf{y}|\mathbf{x})\).

To simplify notation, we define \(\Delta_{\mathrm{BDPO}}\):
\[
   \Delta_{\mathrm{BDPO}} := \beta \left( \log \frac{\pi_{\theta}(\mathbf{y}_{w}|\mathbf{x})}{\pi_{\mathrm{mix}}(\mathbf{y}_{l}|\mathbf{x})} - \log \frac{\pi_{\mathrm{ref}}(\mathbf{y}_{w}|\mathbf{x})}{\pi_{\mathrm{ref}}(\mathbf{y}_{l}|\mathbf{x})} \right).
   \]
Then, we can write \(\mathcal{L}_{\mathrm{BDPO}}=-\mathbb{E}_{\mathcal{D}} \left[ \log \sigma (\Delta_{\mathrm{BDPO}}) \right].\)

Next, we compute the gradient of \(\log \pi_{\mathrm{mix}}(\mathbf{y}_l|\mathbf{x})\) with respect to \(\pi_\theta(\mathbf{y}_l|\mathbf{x})\). Applying the chain rule to the logarithm, we obtain
\[
\nabla_{\pi_\theta(\mathbf{y}_l|\mathbf{x})} \log \pi_{\mathrm{mix}}(\mathbf{y}_l|\mathbf{x}) =+ \frac{\lambda}{\pi_{\mathrm{mix}}(\mathbf{y}_l|\mathbf{x})}.
\]
Using this result, we differentiate \(\Delta_{\mathrm{BDPO}}\) with respect to \(\pi_\theta(\mathbf{y}_l|\mathbf{x})\), yielding
\[
\nabla_{\pi_\theta(\mathbf{y}_l|\mathbf{x})} \Delta_{\mathrm{BDPO}} = -\frac{\beta \lambda}{\pi_{\mathrm{mix}}(\mathbf{y}_l|\mathbf{x})}.
\]
To compute the gradient of the loss function, we apply the chain rule to \(\log \sigma(\Delta_{\mathrm{BDPO}})\), which gives  
\begin{align}
    \nabla_{\pi_\theta(\mathbf{y}_l|\mathbf{x})} \log \sigma(\Delta_{\mathrm{BDPO}}) &= \sigma(-\Delta_{\mathrm{BDPO}}) \cdot \nabla_{\pi_\theta(\mathbf{y}_l|\mathbf{x})}\Delta_{\mathrm{BDPO}} \nonumber\\&=
    -\sigma(-\Delta_{\mathrm{BDPO}}) \cdot \left( \frac{\beta \lambda}{\pi_{\mathrm{mix}}(\mathbf{y}_l|\mathbf{x})} \right)\nonumber .
\end{align}

Finally, taking the expectation over \(\mathcal{D}\), the gradient of \(\mathcal{L}_{\mathrm{BDPO}}\) with respect to \(\pi_\theta(\mathbf{y}_l|\mathbf{x})\) is given by  
\[
\nabla_{\pi_\theta(\mathbf{y}_l|\mathbf{x})} \mathcal{L}_{\mathrm{BDPO}} = \mathbb{E}_{\mathcal{D}} \left[ \beta \cdot \sigma(-\Delta_{\mathrm{BDPO}}) \cdot \frac{\lambda}{\pi_{\mathrm{mix}}(\mathbf{y}_l|\mathbf{x})} \right].
\]

\paragraph{DPO Loss Gradient}
The DPO Loss is:
\[
\mathcal{L}_{\mathrm{DPO}} = -\mathbb{E}_{\mathcal{D}} \left[ \log \sigma \left( \beta \left( \log \frac{\pi_{\theta}(\mathbf{y}_{w}|\mathbf{x})}{\pi_{\mathrm{ref}}(\mathbf{y}_{w}|\mathbf{x})} - \log \frac{\pi_{\theta}(\mathbf{y}_{l}|\mathbf{x})}{\pi_{\mathrm{ref}}(\mathbf{y}_{l}|\mathbf{x})} \right) \right) \right].
\]

we define \(\Delta_{\mathrm{DPO}} := \beta \left( \log \frac{\pi_{\theta}(\mathbf{y}_{w}|\mathbf{x})}{\pi_{\mathrm{ref}}(\mathbf{y}_{w}|\mathbf{x})} - \log \frac{\pi_{\theta}(\mathbf{y}_{l}|\mathbf{x})}{\pi_{\mathrm{ref}}(\mathbf{y}_{l}|\mathbf{x})} \right)\).

Next, we compute the gradient of \(\Delta_{\mathrm{DPO}}\) with respect to \(\pi_\theta(\mathbf{y}_l|\mathbf{x})\). Since only the second term depends on \(\pi_\theta(\mathbf{y}_l|\mathbf{x})\), the gradient is:
\[
\nabla_{\pi_\theta(\mathbf{y}_l|\mathbf{x})} \Delta_{\mathrm{DPO}} = -\frac{\beta}{\pi_\theta(\mathbf{y}_l|\mathbf{x})}.
\]

Similarly BDPO, we obtain:
\[
\nabla_{\pi_\theta(\mathbf{y}_l|\mathbf{x})} \mathcal{L}_{\mathrm{DPO}} = \mathbb{E}_{\mathcal{D}} \left[ \beta \cdot \sigma(-\Delta_{\mathrm{DPO}}) \cdot \frac{1}{\pi_\theta(\mathbf{y}_l|\mathbf{x})} \right].
\]

\paragraph{DPO+NLL Loss Gradient}
The DPO+NLL Loss is:
\[
\mathcal{L}_{\mathrm{DPO+NLL}} = \mathcal{L}_{\mathrm{DPO}} + \alpha \mathcal{L}_{\mathrm{NLL}}, \quad \text{where } \mathcal{L}_{\mathrm{NLL}} = -\mathbb{E}_{\mathcal{D}} \left[ \log \pi_\theta(\mathbf{y}_{w}|\mathbf{x}) \right].
\]

Since \(\mathcal{L}_{\mathrm{NLL}}\) does not contribute to \(\nabla_{\pi_\theta(\mathbf{y}_l|\mathbf{x})}\), the gradient is the same as DPO:
\[
\nabla_{\pi_\theta(\mathbf{y}_l|\mathbf{x})} \mathcal{L}_{\mathrm{DPO+NLL}} = \nabla_{\pi_\theta(\mathbf{y}_l|\mathbf{x})} \mathcal{L}_{\mathrm{DPO}} = \mathbb{E}_{\mathcal{D}} \left[ \beta \cdot \sigma(-\Delta_{\mathrm{DPO}}) \cdot \frac{1}{\pi_\theta(\mathbf{y}_l|\mathbf{x})} \right].
\]

\paragraph{DPOP Loss Gradient}
The DPOP Loss is:
\[
\mathcal{L}_{\mathrm{DPOP}} = -\mathbb{E}_{\mathcal{D}} \left[ \log \sigma \left( \beta \left( \log \frac{\pi_{\theta}(\mathbf{y}_{w}|\mathbf{x})}{\pi_{\mathrm{ref}}(\mathbf{y}_{w}|\mathbf{x})} - \log \frac{\pi_{\theta}(\mathbf{y}_{l}|\mathbf{x})}{\pi_{\mathrm{ref}}(\mathbf{y}_{l}|\mathbf{x})} - \lambda \cdot \max \left( 0, \log \frac{\pi_{\mathrm{ref}}(\mathbf{y}_{w}|\mathbf{x})}{\pi_{\theta}(\mathbf{y}_{w}|\mathbf{x})} \right) \right) \right) \right].
\]

we define \(\Delta_{\mathrm{DPOP}}:= \beta \left( \log \frac{\pi_{\theta}(\mathbf{y}_{w}|\mathbf{x})}{\pi_{\mathrm{ref}}(\mathbf{y}_{w}|\mathbf{x})} - \log \frac{\pi_{\theta}(\mathbf{y}_{l}|\mathbf{x})}{\pi_{\mathrm{ref}}(\mathbf{y}_{l}|\mathbf{x})} - \lambda \cdot \max \left( 0, \log \frac{\pi_{\mathrm{ref}}(\mathbf{y}_{w}|\mathbf{x})}{\pi_{\theta}(\mathbf{y}_{w}|\mathbf{x})} \right) \right)\).

The penalty term \(\max \left( 0, \log \frac{\pi_{\mathrm{ref}}(\mathbf{y}_{w}|\mathbf{x})}{\pi_{\theta}(\mathbf{y}_{w}|\mathbf{x})} \right)\) does not depend on \(\pi_\theta(\mathbf{y}_l|\mathbf{x})\). Thus:
\[
\nabla_{\pi_\theta(\mathbf{y}_l|\mathbf{x})} \Delta_{\mathrm{DPOP}} = -\frac{\beta}{\pi_\theta(\mathbf{y}_l|\mathbf{x})}.
\]

The gradient of \(\mathcal{L}_{\mathrm{DPOP}}\) with respect to \(\pi_\theta(\mathbf{y}_l|\mathbf{x})\) is:
\[
\nabla_{\pi_\theta(\mathbf{y}_l|\mathbf{x})} \mathcal{L}_{\mathrm{DPOP}} = \mathbb{E}_{\mathcal{D}} \left[ \beta \cdot \sigma(-\Delta_{\mathrm{DPOP}}) \cdot \frac{1}{\pi_\theta(\mathbf{y}_l|\mathbf{x})} \right].
\]

\end{document}